\documentclass[journal]{IEEEtran}
\ifCLASSINFOpdf
  \usepackage[pdftex]{graphicx}
\else
\fi
\usepackage{amsmath}
\usepackage{amssymb}
\usepackage{color}

\usepackage[linesnumbered]{algorithm2e}

\DeclareMathOperator*{\argmax}{arg\!\max}
\usepackage{epsfig}
\usepackage{epstopdf}
\usepackage{subcaption}

\begin{document}
%
\title{Game-Theoretic Modeling of Driver and Vehicle Interactions for Verification and Validation of Autonomous Vehicle Control Systems}
%
%
%

\author{Nan~Li,
        Dave~Oyler,
        Mengxuan~Zhang,
        Yildiray~Yildiz,
        Ilya~Kolmanovsky,
        and~Anouck~Girard 
\thanks{N. Li, D. Oyler, M. Zhang, I. Kolmanovsky, and A. Girard are with the Department
of Aerospace Engineering, University of Michigan, Ann Arbor,
MI, 48109 USA.}
\thanks{Y. Yildiz is with the Department of Mechanical
Engineering, Bilkent University, 06800 Cankaya, Ankara, Turkey.}
\thanks{Ilya Kolmanovsky and Nan Li acknowledge the support for this research of the National Science Foundation under Award Number CNS 1544844 to the University of Michigan.}
\thanks{Anouck Girard acknowledges the support for this research of the Air Force Research Laboratory grant FA 8650-07-2-3744 to the University of Michigan.}}

%
%

\markboth{IEEE TRANSACTIONS ON CONTROL SYSTEMS TECHNOLOGY}%
{Shell \MakeLowercase{\textit{et al.}}: Game-Theoretic Modeling of Driver and Vehicle Interactions for Validation and Verification of Autonomous Vehicle Control Systems}
%



\maketitle

\begin{abstract}
Autonomous driving has been the subject of increased interest in recent years both in industry and in academia. Serious efforts are being pursued to address legal, technical and logistical problems and make autonomous cars a viable option for everyday transportation. One significant challenge is the time and effort required for the verification and validation of the decision and control algorithms employed in these vehicles to ensure a safe and comfortable driving experience. Hundreds of thousands of miles of driving tests are required to achieve a well calibrated control system that is capable of operating an autonomous vehicle in an uncertain traffic environment where multiple interactions between vehicles
and drivers simultaneously occur. Traffic simulators where these interactions can be modeled and represented with reasonable fidelity can help decrease the time and effort necessary for the development of the autonomous driving control algorithms by providing a venue where acceptable initial control calibrations can be achieved quickly and safely before actual road tests. In this paper, we present a game-theoretic traffic model that can be used to 1) test and compare various autonomous vehicle decision and control systems and 2) calibrate the parameters of an existing control system. We demonstrate two example case studies, where, in the first case, we test and quantitatively compare two autonomous vehicle control systems in terms of their safety and performance, and, in the second case, we optimize the parameters of an autonomous vehicle control system, utilizing the proposed traffic model and simulation environment.
\end{abstract}


%
\IEEEpeerreviewmaketitle

\section{Introduction}
%
%
%
%
\IEEEPARstart{O}{ne} of the most significant challenges
that must be addressed before autonomous cars can be deployed in mass production is the Verification and Validation (V\&V) of their control systems in terms of safety and performance \cite{Campbell:10, Anderson:16}. It has been estimated that autonomous vehicles need to be driven 275 million miles without fatality to assure the same rate of reliability as existing human driven cars \cite{Kalra:16}. Hence testing and calibration of decision and control systems of autonomous vehicles in simulation becomes necessary to complement the on-the-road tests and the formal methods-based and reachability analysis-based procedures (e.g., see
\cite{wongpiromsarn2008formal,wongpiromsarn2008periodically,lygeros1998verified,althoff2014online}).


One common approach to the design of control systems for autonomous vehicles is to utilize a hierarchical control structure, wherein a higher level outer loop controller generates reference trajectories for the lower level inner loop controller which, in turn, determines the steering angle and acceleration/deceleration inputs required to follow the reference trajectory \cite{Carvalho:15}. Designing these control systems is a challenging task as they need to provide a safe ride together with an acceptable performance and comfort in an uncertain traffic environment. Uncertainties generally emanate from unpredictable behaviors of other drivers, pedestrians, unexpected obstacles, and changing road and weather conditions.  Several control approaches have been proposed for autonomous vehicles including decision trees \cite{Miller:08, Claussman:15}, Partially Observable Markov Decision Processes (POMDPs) \cite{Brechtel:14} and methods based on multi-policy decision making \cite{Galceran:15}, that are mainly employed as outer loop controllers.
For the inner loop, one of the most common approaches is based on Model Predictive Control (MPC) \cite{Falcone:07}-\cite{Carvalho:13}.

Note that advanced driver behavioral models may also be used in the outer loop \cite{Yoo:12, Yoo:13}, with the motivation that an autonomous vehicle should be able to drive at least as well as a human driver. In fact, some experts have suggested that autonomous vehicles should be permitted on public roads only after it is proven that they are superior to human drivers \cite{Anderson:16}.

Simulators can facilitate the development and testing of
autonomous vehicle control algorithms, complementing the road tests.
Since autonomous vehicles will be interacting with human driven vehicles in traffic, such simulators need to reflect driver and vehicle interactions. Several methods have been proposed in the literature to address this problem. In \cite{Lefevre:14} and \cite{Lefevre:15}, a Hidden Markov Model (HMM) based driver model is proposed based on real driving data. In \cite{Vasudevan:12} and \cite{Shia:14}, k-means clustering is used to determine the driving mode and define an approach to predicting and overbounding future vehicle trajectories. It is shown that the performance of an assisted driving algorithm can be improved through the prediction of driver inputs. In \cite{Salvucci:01}, a ``cognitive architectures'' approach, which is ``a computational framework that incorporates built-in, well-tested parameters and constraints on cognitive and perceptual-motor processes,'' is utilized for driver modeling. Built-in logical rules (if-then-else) are used to represent the human decision making process. In \cite{Hidas:02}, lane change behavior of drivers is modeled using a multi agent simulation system called ``Simulation of Intelligent TRAnsport Systems (SITRAS).'' Several logical algorithms are used to model the decision making during lane changes. The resulting actions of the drivers are therefore predefined with strict rules even though driver aggressiveness can be incorporated into the model by tuning certain parameters.

With respect to the existing approaches, the present paper is distinguished by the advanced modeling of driver and vehicle interactions in traffic using a specific game theoretic formulation that is scalable to multiple vehicles. The proposed method has the following advantages: a) actions of drivers and vehicles are determined by utilizing a decision making process, instead of assuming that these actions are prescribed in advance as functions of time or as functions of the state of the system;
b) multiple interactions between human driven vehicles and autonomous vehicles can be modeled simultaneously and in a computationally tractable way;
c) all the vehicles in a traffic scenario are simultaneously modeled as decision makers as opposed to predicting the decisions of one vehicle while assuming that the rest of the vehicles move based on certain kinematic and dynamic constraints.
The proposed game theoretic model makes it possible to conduct a quantitative analysis of the traffic. For example, a) the increase in the number of incidents based on the traffic density can be assessed; b) the effect of a certain parameter value in an autonomous vehicle control algorithm on the safety of the vehicle, e.g., quantified by the number of incidents, can be determined;  c) various autonomous vehicle control algorithms can be compared quantitatively in a multi-vehicle time-extended scenario, based on certain safety and performance metrics; and d) optimization of a cost function that reflects safety and performance can be performed.

Our approach uniquely combines a specific game theoretic formalism, which is used to model intelligent agent interactions, and reinforcement learning, which is used to evolve these interactions in a time-extended (multi-move) scenario. The core ideas are synergistic with the framework of ``semi network-form games,'' \cite{LeeWolpert:11, Backhaus:13} and help us obtain the probable outcomes of a complex traffic scenario driven by multiple interactions. To the authors' knowledge, such an approach has not been previously exploited for automotive traffic modeling.

Other game theoretic approaches, in particular, based on Stackelberg games, have been
studied for application to vehicle highway driving problems in \cite{Yoo:12} and \cite{Yoo:13}. Although these approaches represent driver interactions in traffic using a game theoretic setting, they do not consider dynamic (multi-move) scenarios. The latter are considered in \cite{Ilya:14} for Hybrid Electric Vehicle (HEV) energy management where the driver and the powertrain are considered to be two players in a game. However, increasing the number of players (drivers, in our case) beyond three complicates computing a Stackelberg solution, especially in a time-extended (multi-move) scenario. On the other hand, the hierarchical reasoning based game theoretic approach exploited in this paper is easily scalable. Indeed, an implementation of the proposed approach for a $50$ player game can be found in \cite{Yildiz:14}, and scenarios with up to 30 vehicles are handled in this paper.

To summarize, the contributions of this paper are as follows:
a) we develop a novel and scalable to multiple vehicles traffic model and simulation environment based on a specific game theoretic modeling approach; b) our approach allows the representation of driver interactions for many-move, many-vehicle traffic scenarios using a computationally tractable game theoretic approach, whereas many existing methods consider only one-move interaction between two vehicles;  c) we demonstrate that autonomous vehicle control algorithms can be quantitatively evaluated in our simulator and compared based on safety and performance metrics applied to simulation outcomes; two policies proposed in the literature serve as case studies, while other autonomous driving algorithms could also be easily integrated and tested; d) we provide a case study to demonstrate that probabilistic outcomes of traffic scenarios can be utilized to obtain the optimal calibration for an autonomous vehicle controller.

Preliminary results  have appeared in conference papers \cite{Oyler:16,OylerCDC:16}. Differently from \cite{Oyler:16,OylerCDC:16}, in this paper, a) we incorporate a more realistic action space including harder brakes and faster accelerations; b) we develop a more realistic traffic model with more representative distance constraint violation rates via improvements in the reinforcement learning procedure; and c) we demonstrate that optimal parameter values for an autonomous vehicle control algorithm can be obtained using a cost function based on safety and performance. Furthermore, this paper has additional details and interpretations not found
in our preliminary conference papers.

The paper is organized as follows. In Section~\ref{sec:PD}, we define the problem being treated in this paper and the vehicle model being used. In Section~\ref{sec:DIM}, we present the level-$k$ game theory and the reinforcement learning approach to obtain the policies. In Section~\ref{sec:ADCA}, we introduce two autonomous vehicle control algorithms that will be tested by our simulator. In Section~\ref{sec:Results}, we describe our simulator and its implementation results. We then summarize the key developments in our paper in Section~\ref{sec:Summary}.

\section{Problem Definition} \label{sec:PD}

\begin{figure}[htb]
\centering
 \includegraphics[trim = 0mm 40mm 0mm 30mm, clip, width=8cm]{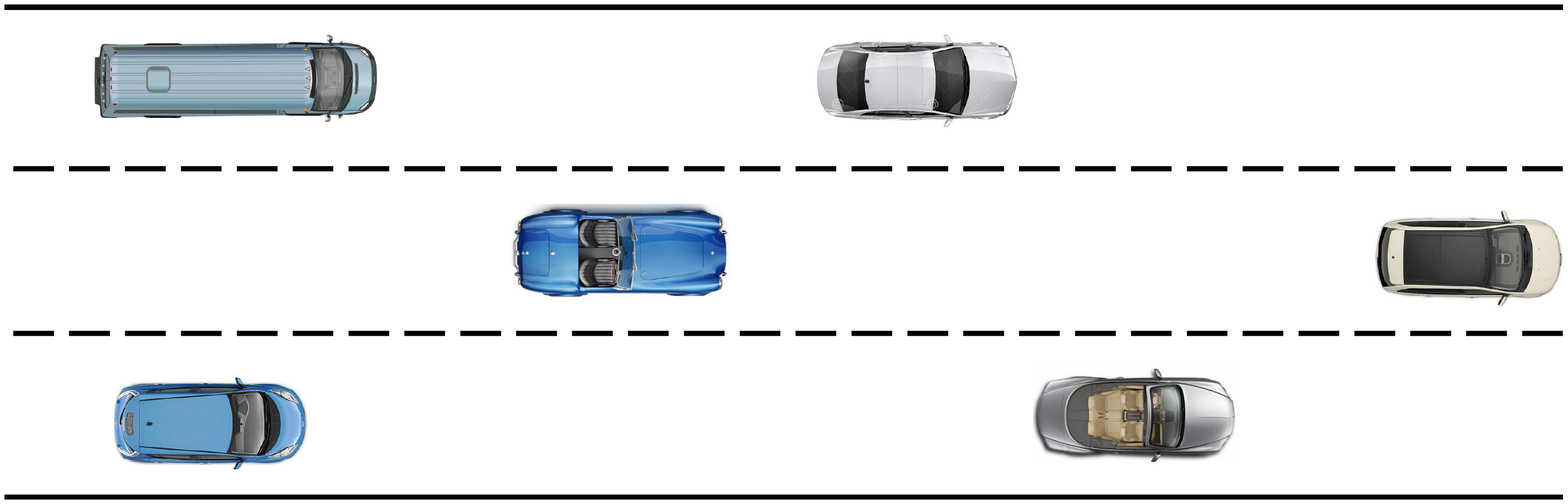}
 \caption{Example scenario: Traffic in a 3-lane highway.}
 \label{f:traffic}
\end{figure}

The problem we treat is to model the behavior of drivers in a traffic scenario where the cars are driven on a 3-lane highway. We later demonstrate that such models can be used in simulators to evaluate autonomous vehicle control policies. Fig.~\ref{f:traffic} shows an example scenario with 6 cars. Note that this is not a restriction of the proposed method, and that scenarios with more cars and lanes can be handled. Simulated cars are assumed to be traveling in the same direction and to be driven by human drivers who obey the general traffic laws.

\subsection{Physical models}

The discrete-time equations of motion for each vehicle during forward motion and lane changes are given by
\begin{align}
x(t+1) &= x(t) + v_x(t)\Delta t, \nonumber \\
v_x(t+1) &= v_x(t) + a(t)\Delta t, \\
y(t+1) &= y(t) + v_y(t) \Delta t, \nonumber
\end{align}
where $x$, $y$ are the vehicle longitudinal and lateral positions, respectively, $v_x$ and $v_y$ are vehicle longitudinal and lateral velocities, respectively, $a(t)$ is the vehicle longitudinal acceleration, and $\Delta t$ is the update period.
The longitudinal acceleration $a(t)$ and the lateral velocity $v_y(t)$ are the two control inputs. We assume that all cars accelerate and decelerate at either $\pm a_1$[m/s$^2$] or $\pm a_2$[m/s$^2$]. The nominal values, $\pm a_1$[m/s$^2$], reflect the acceleration/deceleration a human driver would apply in normal situations, while the hard accelerations/decelerations of $\pm a_2$[m/s$^2$], reflect the values used in urgent situations, based on the maximum acceleration/deceleration capability of a vehicle. We also assume that lane changes occur with constant lateral velocity such that the total time to change lanes is $t_{cl}$[s]. During lane changes, the longitudinal velocity remains constant, and once a lane change begins, it always continues to completion.

While these assumptions represent a reasonable approximation, formulations with more acceleration/deceleration levels can be similarly introduced at the cost of the increased time and effort to compute the policies.

\subsection{Observation space} \label{sec:obs}
In real traffic flow, a driver can neither observe nor process all the information about all cars on the road. A human can possibly observe and use the information he/she obtains from the cars in a certain vicinity of their own. In particular, a human driver can hardly measure his/her exact distances from other cars. He/she can only estimate the distances and specify them as ``close'',``far'', etc. Therefore, we assign the following observation space for the drivers:
\begin{enumerate}
	\item The distance from our car to the car directly in front, called range, quantified as ``close" (distance $\le d_c$[m]),  ``nominal" ($d_c$[m] $<$ distance  $\le d_f$[m]) or ``far" (distance $> d_f$[m]),
	\item The range to the car in the front left lane, quantified as ``close'',  ``nominal'' or ``far,''
	\item The range to the car in the front right lane, quantified as ``close'',  ``nominal'' or ``far,''
	\item The range to the car in the rear left lane, quantified as ``close'',  ``nominal'' or ``far,''
	\item The range to the car in the rear right lane, quantified as ``close'',  ``nominal'' or ``far,''
	\item The range rate to the car in front, quantified as ``approaching'' (distance decreasing),  ``stable'' (distance not changing), or ``moving away'' (distance increasing),
	\item The range rate to the car in the front left lane, quantified as ``approaching'',  ``stable'' or ``moving away,''
	\item The range rate to the car in the front right lane, quantified as ``approaching'',  ``stable'' or ``moving away,''
	\item The range rate to the car in the rear left lane, quantified as ``approaching'',  ``stable'' or ``moving away,''
	\item The range rate to the car in the rear right lane, quantified as ``approaching'',  ``stable'' or ``moving away,''
	\item The lane index of the car, quantified as ``lane 1'',  ``lane 2'' or ``lane 3.''
\end{enumerate}

Note that this description of the observation space reflects the uncertainty (or noise) present in real life driver observation.

\subsection{Action space}
Drivers have 7 basic actions:
\begin{enumerate}
	\item ``Maintain'' current lane and speed,
	\item ``Accelerate'' at rate = $a_1$[$\rm{m/s^2}$], provided velocity does not exceed $v_{\max}$[$\rm{km/h}$],
	\item ``Decelerate'' at rate = $-a_1$[$\rm{m/s^2}$], provided velocity is above $v_{\min}$[$\rm{km/h}$],
	\item ``Hard Accelerate'' at rate = $a_2$[$\rm{m/s^2}$], provided velocity does not exceed $v_{\max}$[$\rm{km/h}$],
	\item ``Hard Decelerate'' at rate = $-a_2$[$\rm{m/s^2}$], provided velocity is above $v_{\min}$[$\rm{km/h}$],
	\item Change lane to the left, provided there is a lane on the left,
	\item Change lane to the right, provided there is a lane on the right.
\end{enumerate}

\subsection{Reward function} \label{sec:RF}

A ``reward function'' is a mathematical representation of the goals of a driver. Basic goals of the drivers in real traffic are 1) to not have an accident, such as a car crash (safety), 2) to minimize the time needed to reach the destination (performance), 3) to keep a reasonable headway from preceding cars (comfort) and 4) to minimize driving effort (comfort).

These goals can be reflected in the reward function given as
\begin{equation}\label{equ:reward_function}
R = w_1 c + w_2 v + w_3 h + w_4 e,
\end{equation}
where $w_i$, $i=1,2,3,4$, are the weights for each term and $c, v, h$ and $e$ represent ``constraint violation,'' ``velocity,'' ``headway'' and ``effort'' metrics. In particular, we define a safe zone for each car (a rectangular area that over-bounds the geometric contour of the car with a safety margin) whose boundaries are treated as distance constraints to represent the safety of the car.

The weighting terms, $w_i$, may change depending on the aggressiveness of the driver, but, intuitively, to avoid distance constraint violation should be of the most importance, i.e.,
\begin{equation}
w_1 \gg w_2, w_3, w_4.
\end{equation}

The terms, $c,v,h,e$, are explained below.

\textbf{$c$ (constraint violation):} The term $c$ is assigned a value of -1 when a distance constraint violation occurs and a value of 0 otherwise.

\textbf{$v$ (velocity):} The term $v$ is assigned the value
\begin{equation}
v = \frac{v_x - v_{\text{nominal}}}{a_1}, \quad v_{\text{nominal}} = \frac{v_{\min}+v_{\max}}{2}.
\end{equation}
Here the reason for dividing by $a_1$ is to make this term of the same order of magnitude as the other terms, to facilitate the choice of weights.

\textbf{$h$ (headway):} The term $h$ takes the following values depending on the headway distance (the range to the car directly in front):
\begin{equation}
h= \left\{
             \begin{array}{lcl}
             -1 &~~\rm{if} &~~\text{headway} \in \text{``close,''} \\
             0 &~~\rm{if} &~~\text{headway} \in \text{``nominal,''} \\
             1 &~~\rm{if} &~~\text{headway} \in \text{``far.''} \\
             \end{array}
        \right.	
\end{equation}

\textbf{$e$ (effort):} The term $e$ takes the value 0 if the driver's action is ``maintain,'' $e_2 = -5$ if the driver's action is ``hard accelerate'' or ``hard decelerate,'' and $e_1 = -1$ otherwise. This term discourages the driver from making unnecessary maneuvers. In particular, a higher penalty discourages the driver from unnecessarily applying ``hard accelerate'' or ``hard decelerate.'' But in the case where another maneuver cannot avoid a constraint violation, the driver would apply ``hard accelerate'' or ``hard decelerate'' to keep safe. Note that the ratio between $e_1$ and $e_2$ depends on the driver's behavior and could be tuned to match the driving behavior of different human drivers.

\subsection{Constraints} \label{sec:Constraints}

The reward function \eqref{equ:reward_function} already reflects the penalty for distance constraint violations, which may be viewed as imposing soft constraints on the control. For some combinations of states and actions that obviously lead to constraint violations, we can also impose hard constraints to avoid the occurrence of such combinations. In particular, we introduce the following hard constraints, which make certain actions unavailable in certain situations:

\begin{enumerate}
	\item If a car in the left lane is in a parallel position, the controlled car cannot change lane to the left,
	\item If a car in the right lane is in a parallel position, the controlled car cannot change lane to the right,
	\item If a car in the left lane is ``close'' and ``approaching,'' the controlled car cannot change lane to the left,
	\item If a car in the right lane is ``close'' and ``approaching,'' the controlled car cannot change lane to the right.
\end{enumerate}
Note that two cars are assumed to be in a parallel position if the safe zones of these two cars intersect in the longitudinal direction. The use of these hard constrains eliminates the clearly undesirable behaviors better than through penalizing them in the reward function, and also increases the learning speed during training.

\section{Driver Interaction Model}\label{sec:DIM}
The driver interaction model developed in this study enables the modeling of driver-driver and driver-autonomous vehicle interactions through the use of hierarchical decision making and reinforcement learning. The model is a ``policy,'' which is a stochastic map from the observation space of the driver to his/her action space (see Section~\ref{sec:PD}). In other words, this map assigns a probability distribution over possible actions for every observation.
Below, we explain how this model is generated.
\subsection{Hierarchical decision making} \label{sec:hdm}
The developed interaction model is premised on the idea that intelligent agents (such as drivers) have different levels of reasoning. According to this observation, a level-0 agent does not consider probable actions of other agents that he/she is interacting with but rather behaves reflexively. For example, if a driver makes a hard brake when he/she observes an obstacle on the road, without considering how the car following behind would react to this sudden deceleration, this behavior is referred to as a level-0 behavior and the driver is referred to as a level-0 driver. On the other hand, if this driver assumes that the car following behind is being driven by a level-0 driver, who would make a hard brake in case of an obstacle, which may not be enough to avoid a collision, and therefore decides to make a lane change to avoid that collision, then this driver is referred to as a level-1 driver. Similarly, if a driver assumes that the other drivers are level-1 and takes an action accordingly, this driver is a level-2 driver. Higher level driver behavior can be modeled using the same logic. A detailed explanation of this hierarchical modeling method is given in \cite{Stahl:95} and \cite{Costa-Gomes:09}.

To obtain higher level policies, one needs to start by defining a level-0 policy. There are various ways to do this, such as selecting each possible action with equal probability regardless of the observation, or constructing a very simple policy, which provides a minimally reasonable behavior for a range of observations.

In this study, a level-0 policy is formulated as follows:

\begin{equation}
\text{action}_{l0} = \begin{cases}
            \text{``Decelerate,''} &\text{if the car in front is }\\
            &\text{``nominal'' and ``approaching,''} \\
            &\text{or is ``close'' and ``stable,''} \\
            \text{``Hard Decelerate,''} &\text{if the car in front is ``close''}\\
            &\text{and ``approaching,''} \\
            \text{``Maintain,''} &\text{otherwise.} \\
\end{cases}	
\end{equation}

\subsection{Reinforcement learning to solve the Partially Observable Markov Decision Process}

The problem treated in this paper is a multi-agent decision making problem. We use a reinforcement learning (RL) algorithm to determine the policies for each agent based on the reward function defined in Section~\ref{sec:RF}. To achieve the maximum reward, the RL algorithm exploits two steps including 1) ``policy evaluation,'' where the state-action pairs are assigned values based on the accumulated rewards they gain, and 2) ``policy improvement,'' where the probability of choosing the actions that have higher reward values are increased. For more details on RL, see \cite{Sutton:98}.

Conventional RL algorithms require the process to be Markov for convergence guarantees. Note that although the underlying dynamics of the highway problem studied in this work is Markov, each agent (driver) can observe only a subspace of the whole state space (see Section~\ref{sec:obs}) and therefore has to solve a Partially Observable Markov Decision Process (POMDP) problem. In RL literature, the observations are commonly referred to as ``messages.'' In this work we employ the Jaakkola RL algorithm \cite{Jaakkola:94}, which distinguishes itself from conventional approaches by guaranteeing to converge at least to a local maximum in terms of average rewards, when the problem is of POMDP type. Below, the Jaakkola RL algorithm is summarized. See \cite{Jaakkola:94} for further details.

The following steps are followed to obtain the driver policies using Jaakkola RL:

\textbf{Step 1.} Evaluate the value function for the messages $V(m|\pi^t)$ and $Q$-values for the message-action pairs $Q(m,a|\pi^t)$ associated with the driver policy $\pi$ at step $t$, using the following equations:
\begin{align}
\beta^t_m(m)=&(1-\frac{\chi^t_m(m)}{K^t_m(m)})\gamma^{(t)}\beta^{t-1}_m(m)+\frac{\chi^t_m(m)}{K^t_m(m)} \notag, \\
V(m|\pi^t)    =& (1-\frac{\chi^t_m(m)}{K^t_m(m)})V(m|\pi^{t-1}) \notag\\
& +\beta^t_m(m)[R^t-\bar{R}(\pi^t)] \notag, \\
\beta^t_a(m,a)=&(1-\frac{\chi^t_a(m,a)}{K^t_a(m,a)})\gamma^{(t)}\beta^{t-1}_a(m,a)+\frac{\chi^t_a(m,a)}{K^t_a(m,a)} \notag, \\
Q(m,a|\pi^t) =& (1-\frac{\chi^t_a(m,a)}{K^t_a(m,a)})Q(m,a|\pi^{t-1}) \notag \\
&+\beta^t_a(m,a)[R^t-\bar{R}(\pi^t)],
\end{align}
where $m$ and $a$ designate ``messages'' (observations) and actions, respectively, the superscript $t$ refers to the time step, while the subscript ($m$ or $a$) indicates whether the function is associated with messages or with message-action pairs. The function $\chi$ is an indicator function, taking the value 1 if the message/message-action pair is visited and 0 otherwise. The function $K$ represents the number of times a particular message/message-action pair is visited. The positive sequence $\gamma^{(t)}$ represents a discount factor and is converging to 1 in the limit as $t \rightarrow \infty$. The $R^t$ is the reward obtained at the current time instant $t$, while $\bar{R}(\pi^t)$ is the average reward associated with the policy $\pi^t$, i.e., if this policy were executed for an infinite time \cite{mahadevan1996average}.

In the implementation of the above algorithm, the true average reward $\bar{R}(\pi^t)$ is replaced with an estimate, which is computed as an average reward over a past window, making use of the fact that the policy is slowly varying in time.

%

\textbf{Step 2.} Update the driver policy $\pi$, which is a probabilistic mapping from observations (messages) to actions, using the following equation:
\begin{equation}
\pi^{t+1}(a|m) = (1-\varepsilon)\pi^t(a|m)+\varepsilon \hat{\pi}^{t}(a|m),
\end{equation}
where $0<\varepsilon<1$ is the learning rate and $\hat{\pi}^{t}$ is chosen such that $J(m|\hat{\pi}^{t})=\max\limits_{a} \big(Q(m,a|\pi^t)-V(m|\pi^t)\big)$. For any policy $\pi(a|m)$, $J(m|\pi)$ is defined as:
\begin{equation}
J(m|\pi)= \sum\nolimits_{a} \pi(a|m)\big(Q(m,a|\pi)-V(m|\pi)\big).
\end{equation}
Note that based on the process defined above, $\hat{\pi}^t$ should be defined in such a way that it has probability 1 for picking the action $a$ that has the highest $Q(m,a|\pi^t)$ value and probability 0 for picking the
other actions.

Going through step 1 and 2 at each time $t$, the driver model, or the converged policy, denoted by $\pi^*$, is obtained once the policy converges during the iterative process. The convergence criterion is based on the convergence of the average reward, i.e., change in the average reward within a given number of steps less than a tolerance in absolute value.

\subsection{The role of hierarchical decision making in obtaining driver policies}

The process of obtaining driver policies is called ``training,'' where the trained driver is a \textit{learner} and the other vehicles and automation constitute the \textit{environment}. During the training process, the model of the environment is needed to obtain state transitions as a result of driver actions, where the hierarchical decision making approach plays a crucial rule: for the training of a level-$k$ driver policy, all of the traffic but the trained driver are assigned level-($k$-1) policies. The process starts with the determination of a level-0 policy (see Section~\ref{sec:hdm}), which represents the lowest level where the drivers do not consider interaction with other drivers and do not explicitly take into account their possible actions. Once a level-0 policy is determined, the RL algorithm is run by assigning level-0 policies to all of the vehicles except the one that is being trained. At the end of the training process, a level-1 policy is obtained. Similarly, while training a level-2 policy, all of the vehicles but the trained vehicle are assigned a level-1 policy. This hierarchical assignment continues until the desired highest level is obtained. In experimental studies \cite{Costa-Gomes:09}, it is shown that in human interactions, level-3 players are very rarely encountered and therefore in our results we trained policies up to and including level-2.
\section{Autonomous Driving Control Approaches}\label{sec:ADCA}
The proposed traffic model has been employed to build a simulator to test and evaluate the performance of autonomous driving control algorithms. As specific examples, two autonomous driving approaches, based on Stackelberg policies and decision trees, will be evaluated and compared, using a simulator in which the traffic, other than the controlled vehicle, consists of drivers modeled using our game theoretic policies. In this section, the control algorithms that will be tested are briefly explained and in the next section simulation evaluations are provided.

\begin{figure}[h!]
\begin{center}
\begin{picture}(80.0, 90.0)

\put(  5,  0){\epsfig{file=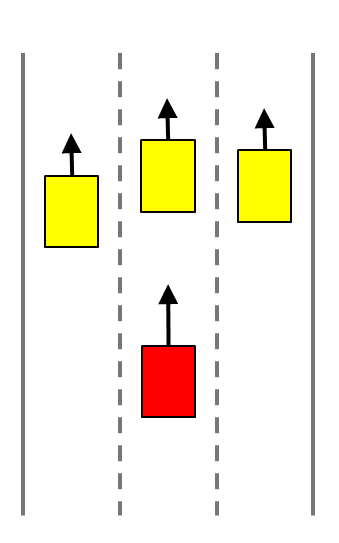,height=1.4in}}  

\end{picture}
\end{center}
      \caption{No solution for the red car by only changing lanes.}
      \label{fig:case_noSoln}
\end{figure}

The Stackelberg policies and the decision tree policies that are compared in this study were originally developed in \cite{Claussman:15}, \cite{Yoo:12} and \cite{Yoo:13}. Since these policies were developed under assumptions representing a simpler traffic environment, some necessary modifications were made to let the autonomous vehicle, which will employ these policies, be able to operate in the traffic environment investigated in this study. For example, the originally proposed Stackelberg and decision tree policies consider only lane change actions. To make them more compatible with the test environment, acceleration/deceleration actions are added to their action space. Fig.~\ref{fig:case_noSoln} shows the necessity of this modification, where each rectangle represents a car and the arrows represent both the driving direction and velocity (the longer the arrow, the larger the velocity). In the figure, three yellow cars are in front of the red test car and since the speed of the test car is larger than that of the yellow cars, there is a danger of collision that cannot be resolved only with a lane change.

\subsection{Stackelberg policies}

To generate Stackelberg policies for the autonomous vehicle, we consider three vehicles as players, and the rest of the vehicles are considered to be the environment. The three players are assigned roles as the ``leader,'' ``first follower,'' and ``second follower,'' and they choose their actions sequentially: the leader chooses its action first, followed by the first follower, and, finally, the second follower. Each player evaluates their actions according to a utility function that consists of two parts. The first part, known as the positive utility, is defined as follows:
\begin{equation}
	U_{\rm pos} = \begin{cases}
			\min(d_{\triangle},d_v), &\text{if there is a vehicle ahead},\\
			d_v, &\text{otherwise},
			\end{cases}
\end{equation}
where $d_{\triangle}$ is the distance to the car directly in front, i.e., the headway distance, and $d_v$ is the maximum visibility distance. The second part of the utility is known as the negative utility:
\begin{equation}
	U_{\rm neg} = d_{\nabla}-v_rT - d_{\min},
\end{equation}
where $d_{\nabla}$ and $v_r$ are the distance to and the relative velocity of the car immediately behind, $T$ is a prediction time window, and $d_{\min}$ is the minimum distance required to allow a lane change; here, $d_{\min}$ is set to the car safe zone length. Thus, overtaking vehicles are taken into consideration, and lane changes that cut off overtaking vehicles are discouraged.

The actions chosen by the leader, first follower, and second follower, denoted by $\gamma_\ell$, $\gamma_{f1}$, and $\gamma_{f2}$, respectively, are the Stackelberg Equilibrium actions; i.e., the leader chooses its actions to maximize its utility for the worst-case actions that the two followers might choose. Thus, the leader chooses:
\begin{equation}
	\gamma_\ell^*=\max_{\gamma_\ell} \min_{\gamma_{f1}} \min_{\gamma_{f2}} \biggl[ U'_{\rm pos} + U'_{\rm neg}\biggr],
\end{equation}
where $U'_{\rm pos}$ and $U'_{\rm neg}$ are the utilities that correspond to a specific set of actions $\{\gamma_\ell,\gamma_{f1},\gamma_{f2}\}$. The two followers maximize their own utilities with the known choice of $\gamma_\ell$. In this paper, when constructing Stackelberg policies, the controlled vehicle is the leader, and the two cars immediately behind are the followers. An alternative choice in which the controlled vehicle is one of the followers instead of the leader can be treated similarly.

It is noted that in \cite{Yoo:12} and \cite{Yoo:13} vehicle dynamics are different than the ones used in this paper. Furthermore, some aspects of the modeling such as uncertainties in measuring distance, side-viewing and response delays that are considered in these references are omitted here to simplify the analysis but can be easily integrated.

\subsection{Decision tree policies}

In the decision tree approach to autonomous driving, the vehicle's actions are determined by a path planner that evaluates a specified number of pre-selected action profiles by building a tree of potential action sequences and evaluating each branch according to a specified metric.

\begin{figure}[h!]
\begin{center}
\begin{picture}(190.0, 172.0)
\put(  0,  0){\epsfig{file=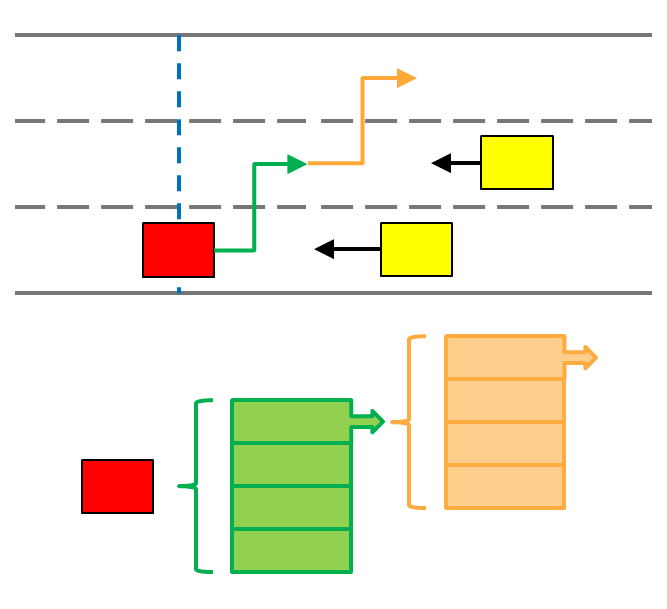,width=2.65in}}  
\small
\put(  69,51.5){To left}
\put(  71,39){Accel.}
\put(  82,25){\vdots}
\put(  69,14){To right}
\put(  130,70){To left}
\put(  132,57){Accel.}
\put(  143,43){\vdots}
\put(  130,32){To right}
\normalsize
\put(65,67){1st layer}
\put(125,15){2nd layer}

\put(49,170){$0 \rightarrow x$}

\end{picture}
\end{center}
      \caption{Decision tree diagram. The black arrows show the relative velocities of the yellow cars with respect to the red test car.}
      \label{fig:DT_diagram}
\end{figure}
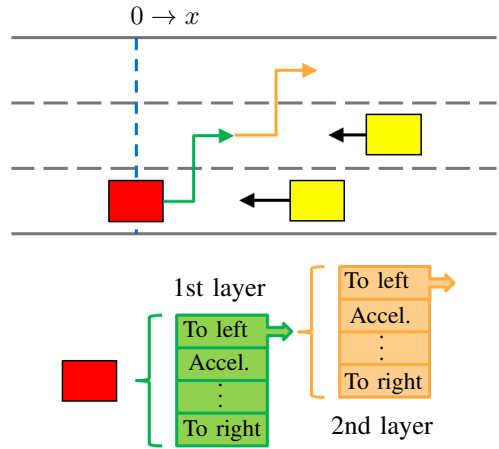

In this paper, the decision tree consists of two layers, where each layer allows the seven actions listed in Section~\ref{sec:PD}-C. Thus, the action profiles consist of two actions each, and 49 profiles are evaluated. The evaluation metric is based on the reward function \eqref{equ:reward_function}, which is also used for the training of the level-$k$ policies. In particular, the total reward is calculated as a weighted sum of the rewards obtained from the two layers:
\begin{equation} \label{eq:Rtot}
R_{\rm total} = w_{l1} R_{l1} + w_{l2} R_{l2},
\end{equation}
where $w_{l1}, w_{l2} \in \mathbb{R}^+$ are the weighting terms. The car applies the first-layer action of the profile that has the highest total reward among all profiles, and repeats this procedure at each step. When evaluating the action profiles, the controlled vehicle assumes that all other vehicles would apply the action ``maintain'' during the prediction horizon.

In \cite{Claussman:15}, it was assumed that the environment evolves deterministically over a planning horizon, independently of the controlled vehicle's actions. However, in our simulator, not only the controlled vehicle responds to the traffic, but also the traffic responds to the controlled vehicle's actions.

\subsection{Path planner triggering threshold}

Both the Stackelberg and the decision tree policies are used as path planners and triggered only when necessary and beneficial. When the policies are not triggered, the vehicle follows a predefined driving pattern. More specifically, with the help of Fig.~\ref{fig:trigger}, the driving algorithm can be explained as follows:

\begin{figure}[h!]
\begin{center}
\begin{picture}(170.0, 170.0)
\put(  0,  0){\epsfig{file=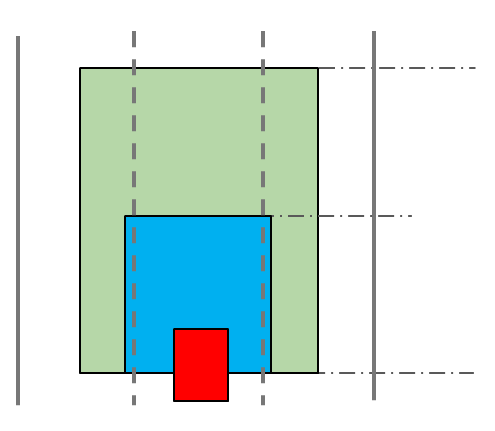,width=2.7in}}  
\put(  72,23){Car}
\put(  25,150){left}
\put(  66,150){middle}
\put(  118,150){right}
\put(76,110){A}
\put(76,55){B}
\put(166,82){$x_A=42$[m]}
\put(156,53){$x_B=21$[m]}

\end{picture}
\end{center}
      \caption{Triggering threshold.}
      \label{fig:trigger}
\end{figure}
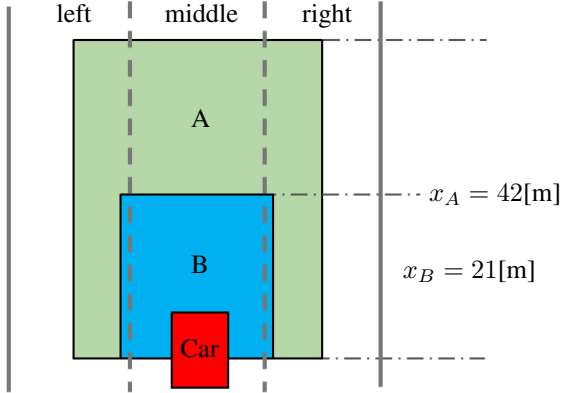

\begin{enumerate}
	\item ``Accelerate,'' if there are no other cars in region A,
	\item Path planner is triggered, if there are cars in region A, but no cars in region B,
	\item ``Safe Mode,'' which in this paper is the same as the level-0 policy, is applied if there are cars in region B.
\end{enumerate}

It is clear that if there are no cars in region A, the autonomous vehicle may accelerate safely, while if there are cars in region B, the autonomous vehicle is supposed to decelerate to keep a reasonable headway distance. The activation logic is designed 1) to increase the safety, and 2) to reduce the computational cost by preventing unnecessary action evaluations. In particular, region A is designed to cover the center of adjacent lanes, while region B is designed to cover the boundary lines of the current lane, so that when some vehicle in the vicinity is changing lanes into the test vehicle's lane, the test vehicle becomes aware of it before the other vehicle actually enters its lane.

\section{Results}\label{sec:Results}

\subsection{Environment and set-up}

We model the environment as follows. The width of a lane is 3.6[m], and the safe zones around the cars, which should not be violated, are modeled as 6[m] $\times$ 2[m] boxes. Cars always drive at the center of a lane unless they are changing lanes. Cars only accelerate or decelerate at $\pm a_1 = \pm 2.5$[m/s$^2$] or $\pm a_2 = \pm 5$[m/s$^2$], and lane changes occur with constant lateral velocity such that the total time to change lanes is $t_{cl} = 2$[s]. During lane changes, the longitudinal velocity remains constant and once a lane change begins, it always continues to completion. The longitudinal axis is called $x$, and its origin is collocated with the car that is to be trained or evaluated. 

To configure the simulation, the following values need to be specified:
\begin{enumerate}
	\item the number of lanes, $n_\ell$,
	\item the number of cars, $n_c$,
	\item the maximum allowable initialization distance, $x^0_{\max}$,
	\item the simulation duration, $t_f$.
\end{enumerate}

The following procedure is employed to initialize a simulation: A car is assigned to a lane that is determined randomly based on the uniform distribution. The specific location of the car within the assigned lane is determined randomly based on the uniform distribution in $[-x^0_{\max},x^0_{\max}]$. Then, the initial longitudinal velocity of the car is given randomly based
on the uniform distribution within the range $[v_{\min},v_{\max}]=[62,98]$[$\rm{km/h}$]. The initial action is set as ``maintain.'' The car is then assigned a policy to follow (level-0, 1, or 2). This process is repeated until all cars are configured. While locating each car, it is required that the minimum initial distance between the cars be 30[m]. Once the initialization is completed, the simulation is run according to Algorithm \ref{alg:Episode} given below.

\begin{algorithm}[h]
	$t=0$.\\
	\While{$t<t_f$}{
		\ForEach{car}{
			Obtain observations from the environment.\\
            Given the observations, determine an action based on assigned policy. \\
			Given the action, update position and velocity.\\
		}
		\If{training a policy}{
			Evaluate reward function for trainee.\label{algline:reward}\\
			Update value function.\label{algline:Jaakkola}
		}
		\If{trainee/test vehicle is in a constraint violation state}{End the simulation.}
		$t = t + \Delta t$.
	}
\caption{Single Episode Simulation}
\label{alg:Episode}
\end{algorithm}

Five cars are observable by the test vehicle, as described in Section~\ref{sec:obs}, and a car is considered ``close'' if its relative longitudinal position satisfies $|x_r| \le d_c=21$[m], ``nominal" if $d_c< |x_r| \le d_f=42$[m], and ``far'' if $ d_f< |x_r| \le d_v=63$[m], where $d_v$ is the maximum visibility distance. Cars farther away than $d_v$[m] are considered to be out of visual range and unobservable.  If no car can be observed in a position, this is considered equivalent to a car that is ``far'' and ``moving away.'' Note that $d_c = 21$[m] is determined by considering the minimum distance needed for a car to safely avoid a distance constraint violation by braking with the maximum allowable deceleration in the worst case scenario where a car in front is approaching with maximum relative speed and entering the ``close'' range.

\begin{figure}[h!]
\begin{center}
\begin{picture}(220.0, 100.0)
\put(  0,  0){\epsfig{file=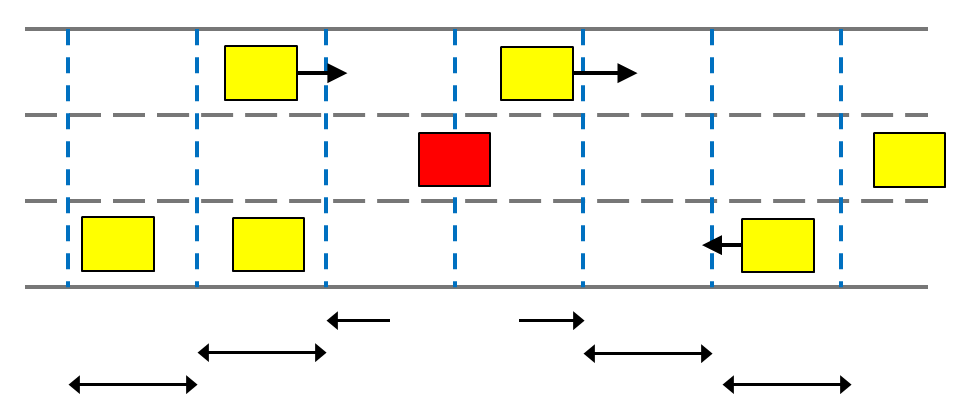,width=3in}}  
\put(  92.5,18.75){close}
\put(  88.5,10.5){nominal}
\put(  97.0,2.25){far}
\put(0,75){3}
\put(0,55){2}
\put(0,35){1}
\put(100,95){$0 \rightarrow x$}

\end{picture}
\end{center}
      \caption{Simulation environment.}
      \label{fig:environment}
\end{figure}
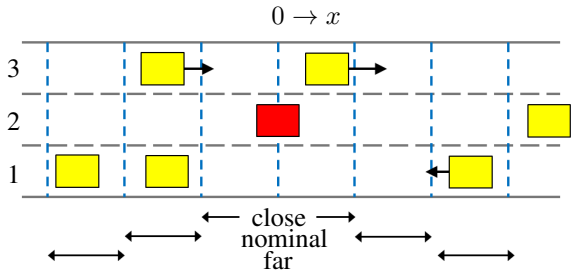

Fig.~\ref{fig:environment} shows a snapshot of an example simulation setup with three lanes. The rectangles represent the safe zones around the cars, and the arrows attached to the rectangles show the relative velocities of the cars with respect to the test vehicle, that is located in the center lane at $x=0$. Specific observations by the test vehicle are given as follows:
\begin{itemize}
	\item Front left: close, moving away,
	\item Front center: far, moving away,
	\item Front right: far, approaching,
	\item Rear left: nominal, approaching,
	\item Rear right: nominal, stable,
	\item Lane index: lane 2.
\end{itemize}

Note that two cars in Fig.~\ref{fig:environment} are unobservable: The car in the front center position is beyond the visual range -- so the corresponding observed status is ``moving away'' even though the car is actually ``stable;'' and the car in the rear right ``far'' position is hidden by the car in the rear right ``nominal'' position. These limitations in the observation space reflect the POMDP nature of the problem, as discussed previously.

\subsection{Level-0 driver behavior}
In this section, we present simulation results to show the driving behavior of a level-0 car. Furthermore, we present the simulator user interface.

\begin{figure}[h!]
\begin{center}
\begin{picture}(220.0, 300.0)

\put(  0,  200){\epsfig{file=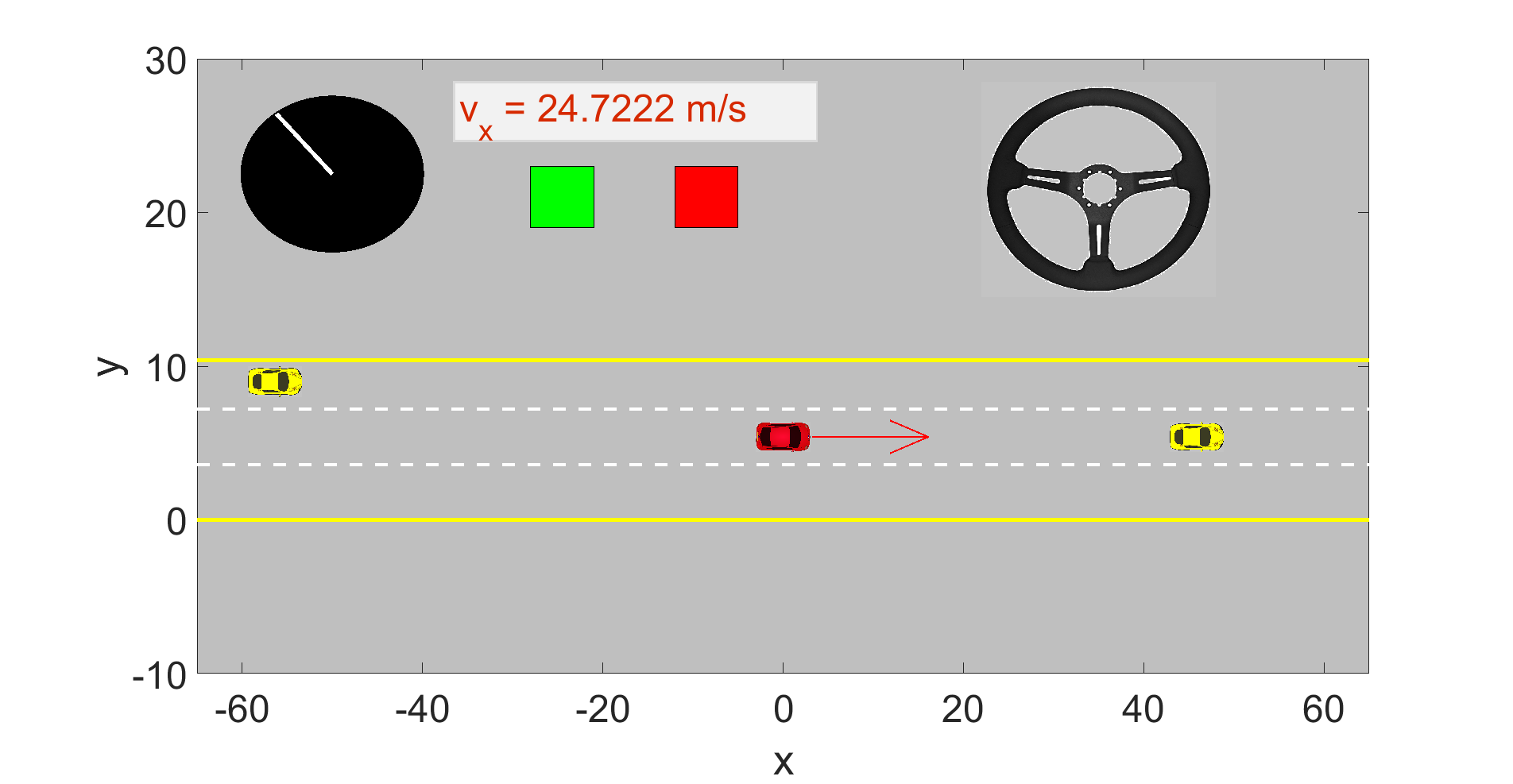,width=2.7in}}  
\put(  0,  100.0){\epsfig{file=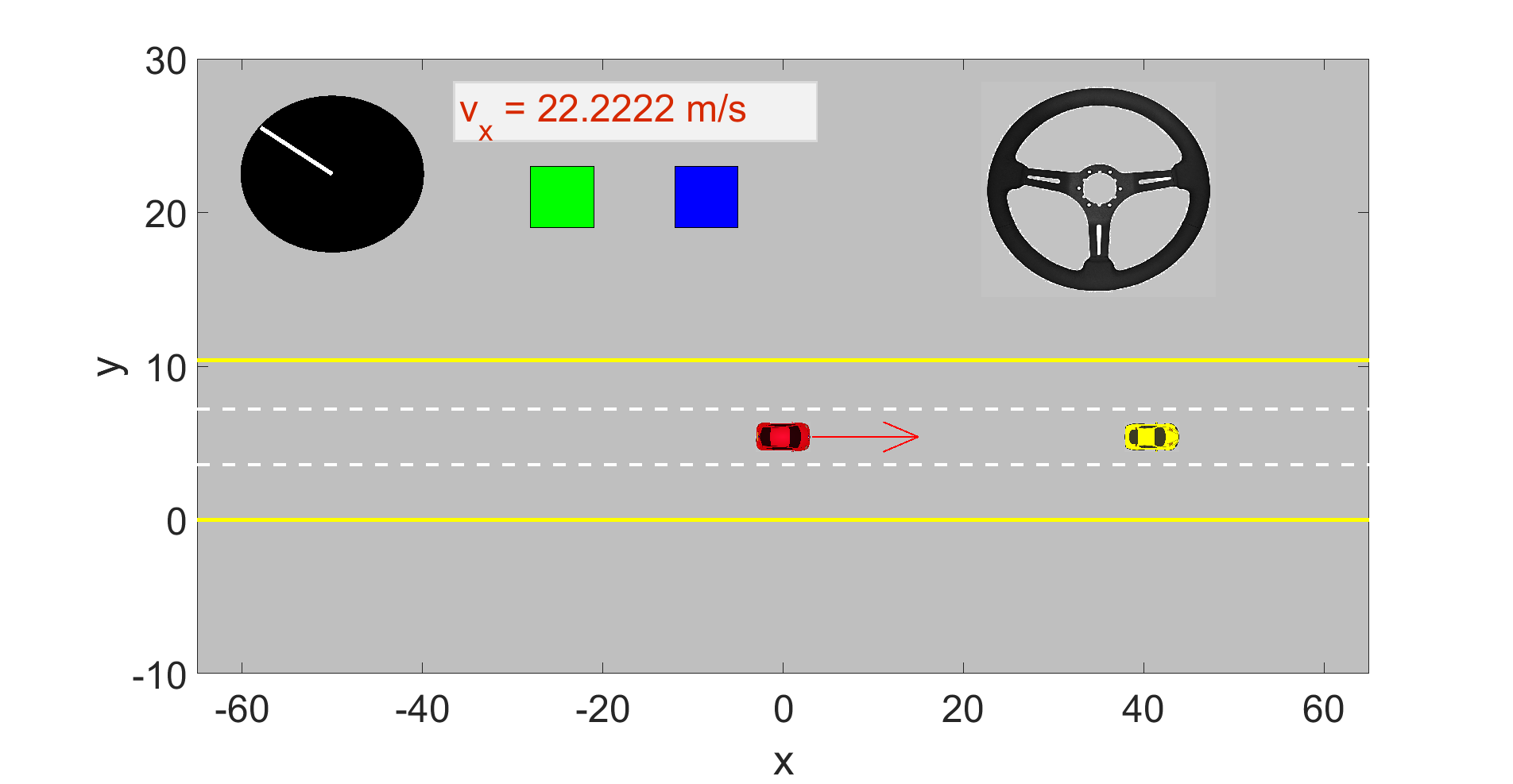,width=2.7in}}  
\put(  0,  0){\epsfig{file=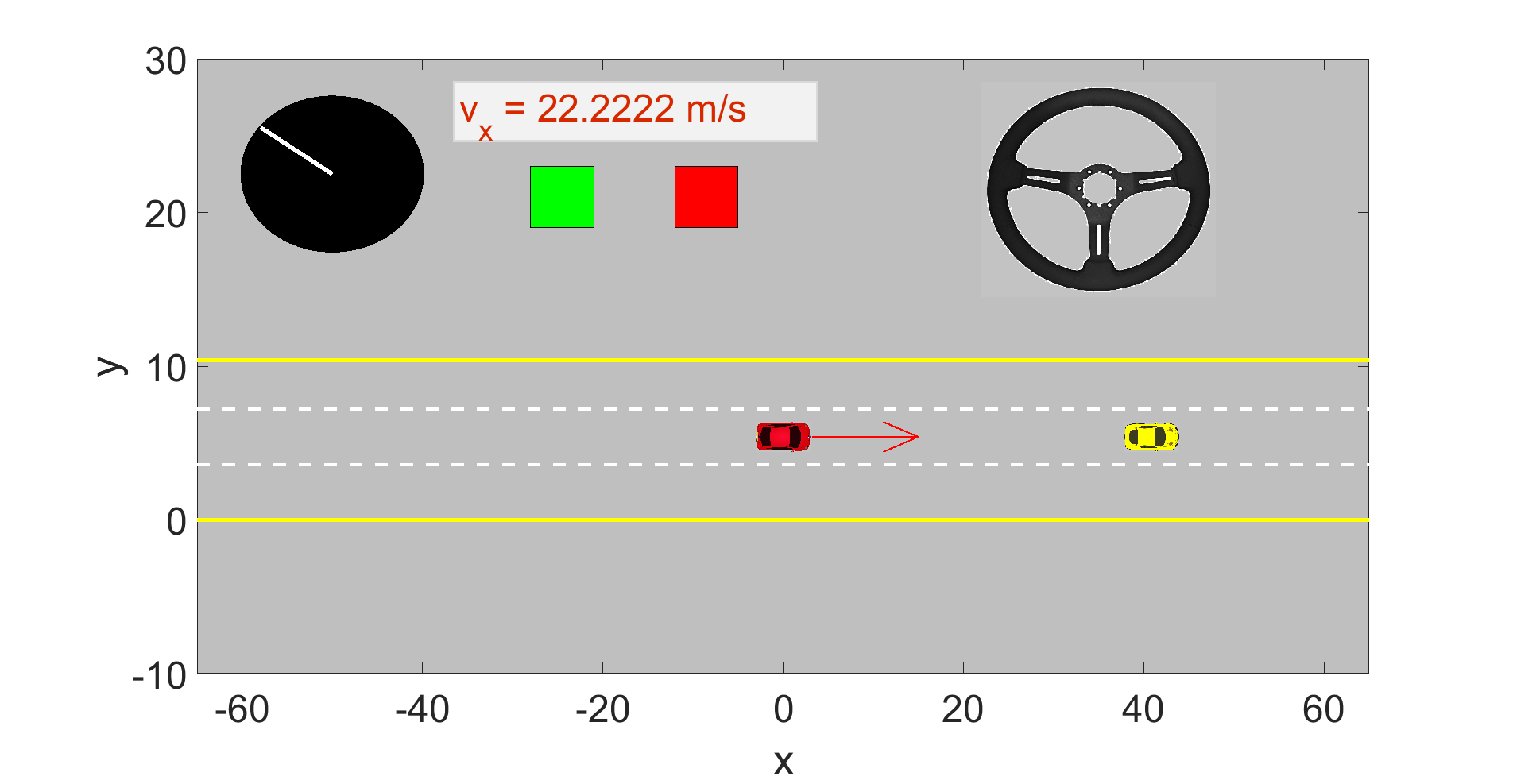,width=2.7in}}  

\put(  186,  280.0){(a)}  
\put(  186,  265.0){40[s]}  
\put(  186,  180.0){(b)}  
\put(  186,  165.0){42[s]}  
\put(  186,  80.0){(c)}  
\put(  186,  65.0){44[s]}  

\end{picture}
\end{center}
      \caption{Level-0 simulation results: Plots (a)-(c) show snapshots of the simulation at 40[s], 42[s] and 44[s], respectively.}
      \label{fig:Sim_Level0}
\end{figure}
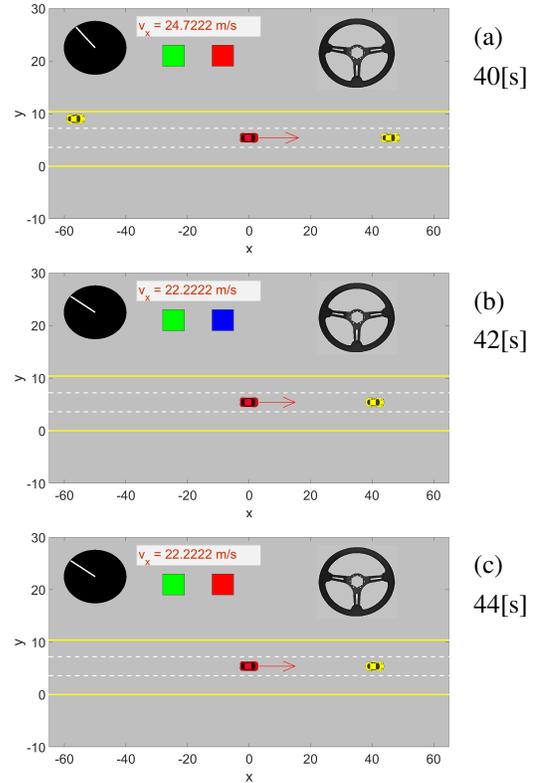

In Fig.~\ref{fig:Sim_Level0}, the red car in the center is the trainee/test vehicle, while the yellow cars make up the traffic environment. The red arrow in front of the red car indicates its travel direction, and arrow size indicates how fast the car is traveling. The panel on the left is a speedometer, and the steering wheel on the right indicates the lateral motion of the car. The green box and red box in the middle indicate the gas pedal and the brake pedal, respectively, and when any of them turns blue, that indicates the pedal is pressed. The coordinate axis is fixed on the test car and the motions of the other cars can be tracked by their relative distance to the red car. In Fig.~\ref{fig:Sim_Level0}(a), a yellow car directly in front of the red car (``far'') is ``approaching'' because the red car is faster. At this moment neither the gas pedal nor the brake pedal is pressed. In Fig.~\ref{fig:Sim_Level0}(b), the yellow car enters the ``nominal'' range, and consistently with the level-0 policy, the red car brakes and decreases its speed. In Fig.~\ref{fig:Sim_Level0}(c), the red car gets to a lower speed, and the yellow car is now ``stable.''

\subsection{Training process}

When training a new policy, the observation value function, $V$, for observed message $m$, and the action value function, $Q$, for message-action pair $(m,a)$, are initialized as follows:
\begin{equation}
\begin{aligned}
	&\forall m, ~~V(m)=0;\\
	&\forall m, \forall a, ~~ Q(m,a)=0.
\end{aligned}
\end{equation}
For each observation, the actions are assigned equal probability of selection at initialization, and during each policy improvement step, if
\begin{equation}
\max_a Q(m,a)>V(m),
\end{equation}
then 0.01 is added to the probability of selecting $\argmax_a Q(m,a)$, after which the action probabilities are normalized.

The observation space described in Section~\ref{sec:obs} has $3^{11}$ different observations. In order to ensure that the learning algorithm is exposed to a large portion of the observation space, the trainee needs to be exposed to both sparse and dense traffic environments. Therefore, during training, the number of cars in the environment is selected randomly, based on the uniform distribution, where $0\leq n_c \leq n_c^{\max}$. The maximum number of cars, $n_c^{\max}$, is chosen based on the number of lanes and $x^0_{\max}$, such that if $n_c^{\max}$ cars are placed in the environment, the road is near full capacity.

Finally, after sufficient training time, the level-0 policy is assigned to the observations (messages) that are still not visited enough during training, so that conservative actions are performed in such rarely encountered cases.

Training then proceeds according to Algorithm \ref{alg:Training}, given below:
\begin{algorithm}
	step=0;\\
	\While{step $<$ desired training cycles}{
		Randomly select $n_c\in [0,n_c^{\max}]$.\\
		Initialize all cars with level-($k$-1) policies.\\
		Evaluate the level-$k$ policy using Algorithm \ref{alg:Episode}.\\
		Improve the policy.\\
		step=step+1. \\
	}
	Assign level-0 policy to the not well trained cases.
\caption{Training Process}
\label{alg:Training}
\end{algorithm}

\begin{figure}[h!]
\begin{center}
\begin{picture}(245.0, 120.0)

\put(  -10,  0){\epsfig{file=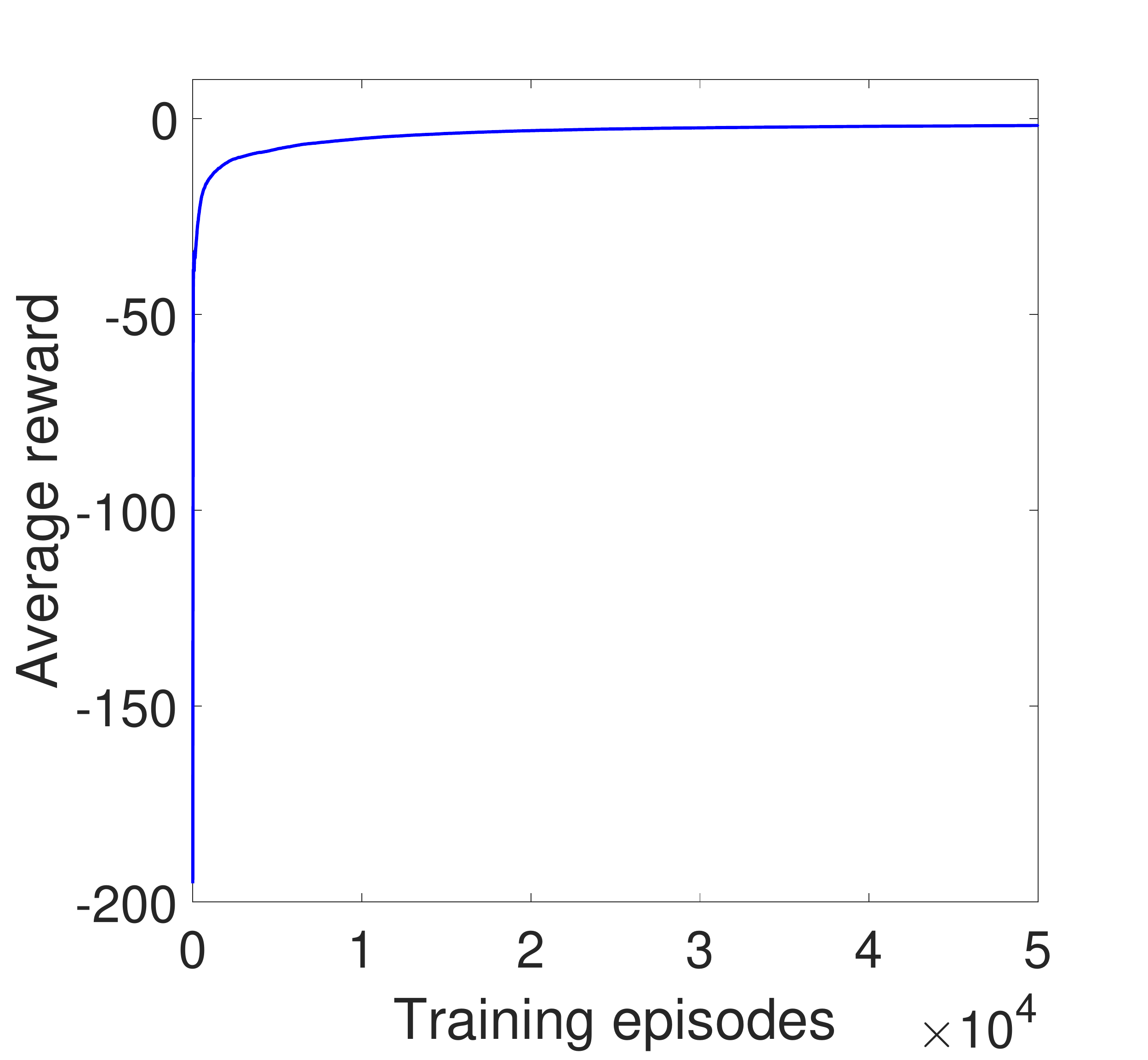,width=1.9in}}  
\put(  120, 0){\epsfig{file=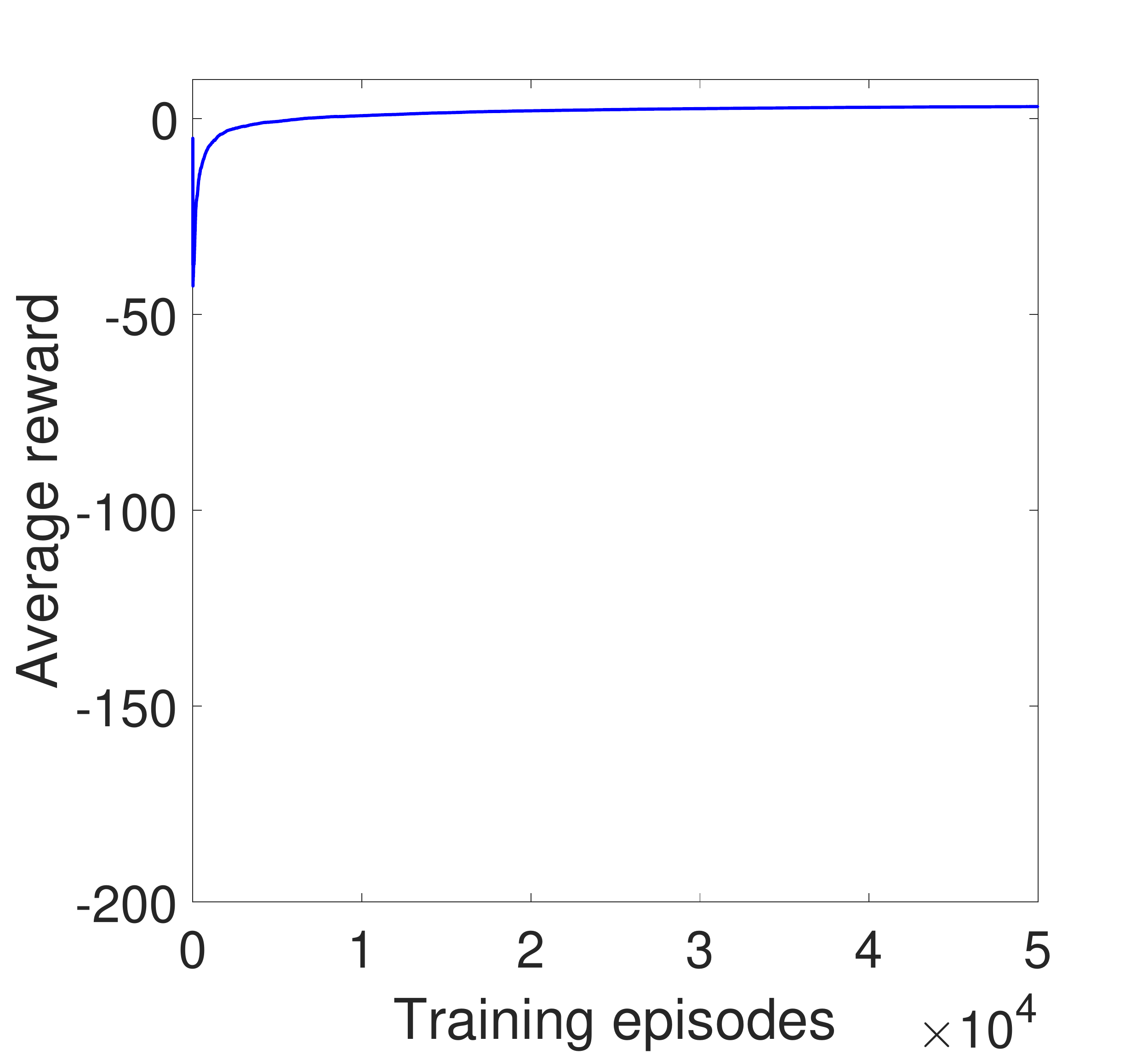,width=1.9in}}  
\put(  95,  95){(a)}  
\put(  225,  95){(b)}  

\end{picture}
\end{center}
      \caption{Average rewards: (a) Training for level-1. (b) Training for level-2.}
      \label{fig:rewards}
\end{figure}

Fig.~\ref{fig:rewards} shows the evolution of the average reward during level-1 and level-2 trainings. The reward function weights are chosen as: $w_1 = 10000$, $w_2 = 5$, $w_3 = w_4 = 1$.

\subsection{Level-$k$ driver interactions}

\begin{figure}[h!]
\begin{center}
\begin{picture}(220.0, 600.0)




\put(  0,  500){\epsfig{file=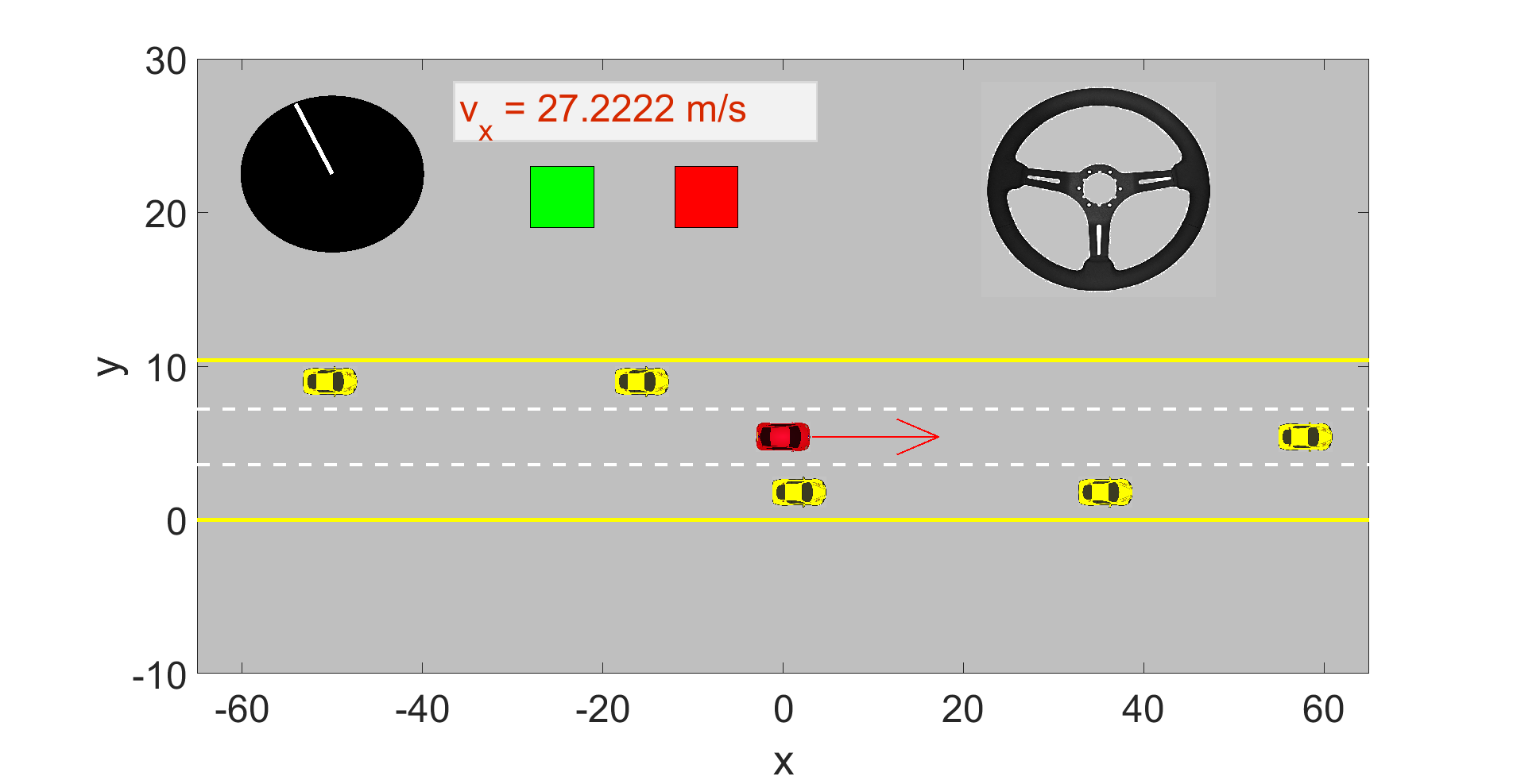,width=2.7in}}  
\put(  0,  400.0){\epsfig{file=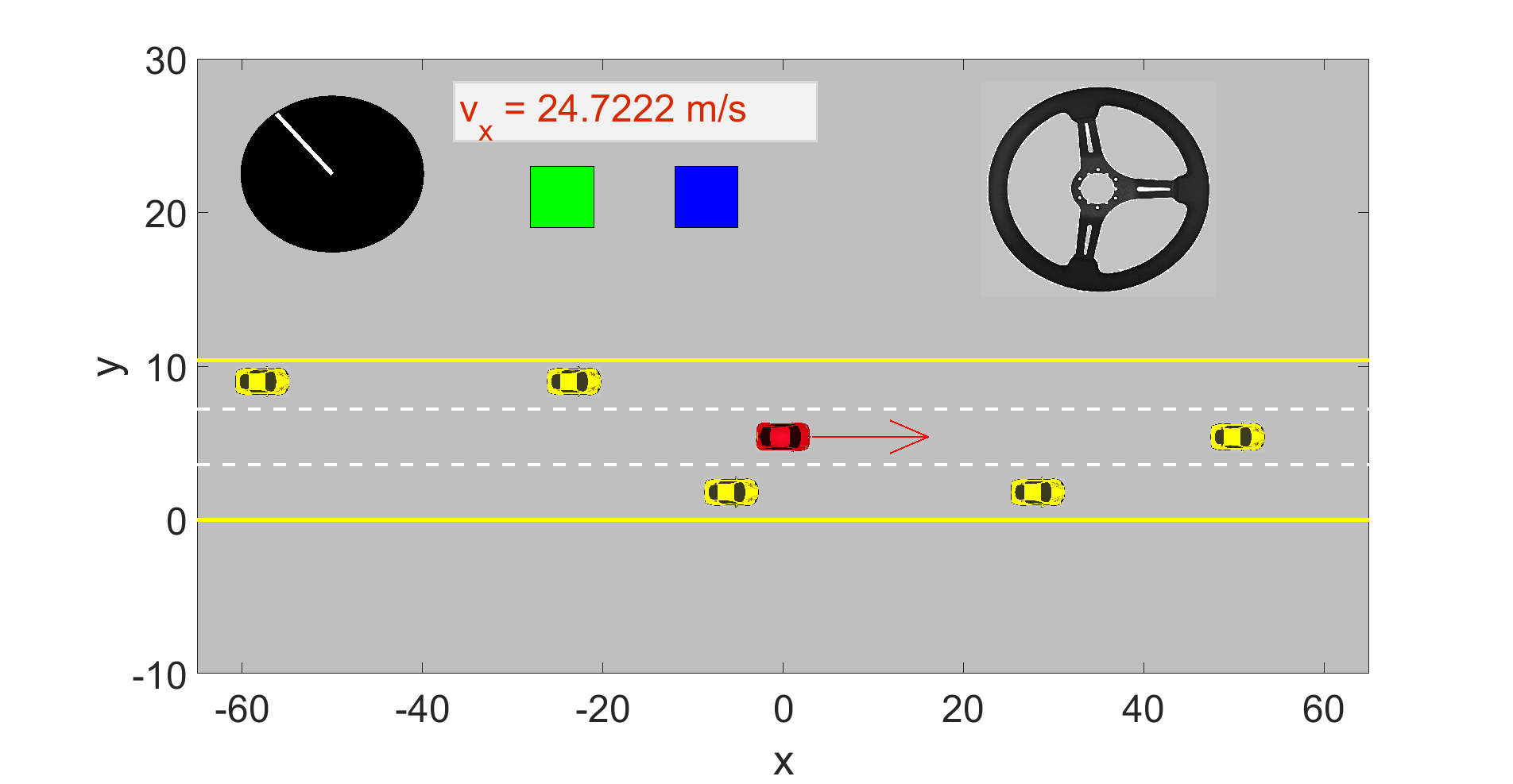,width=2.7in}}  
\put(  0,  300){\epsfig{file=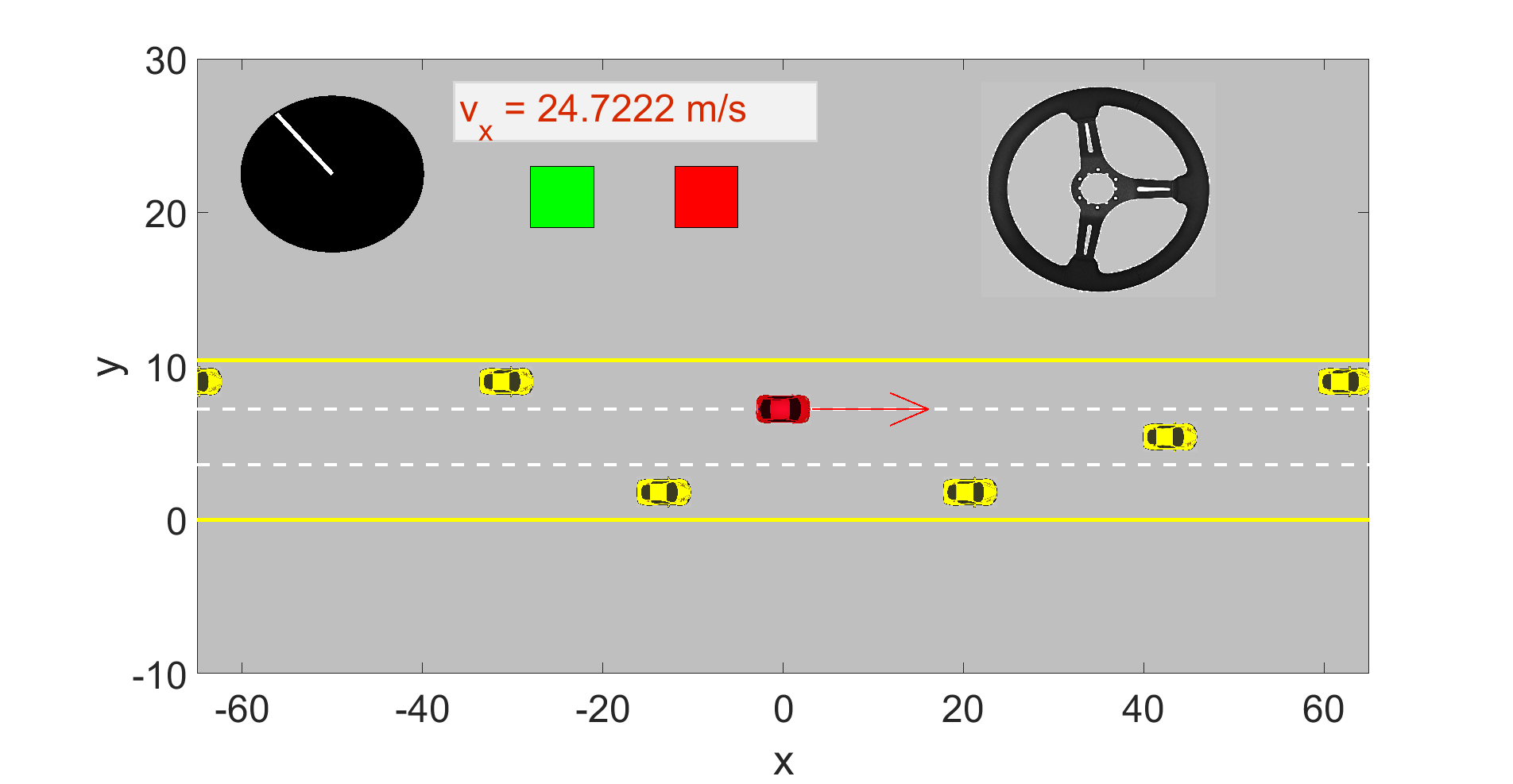,width=2.7in}}  
\put(  0,  200){\epsfig{file=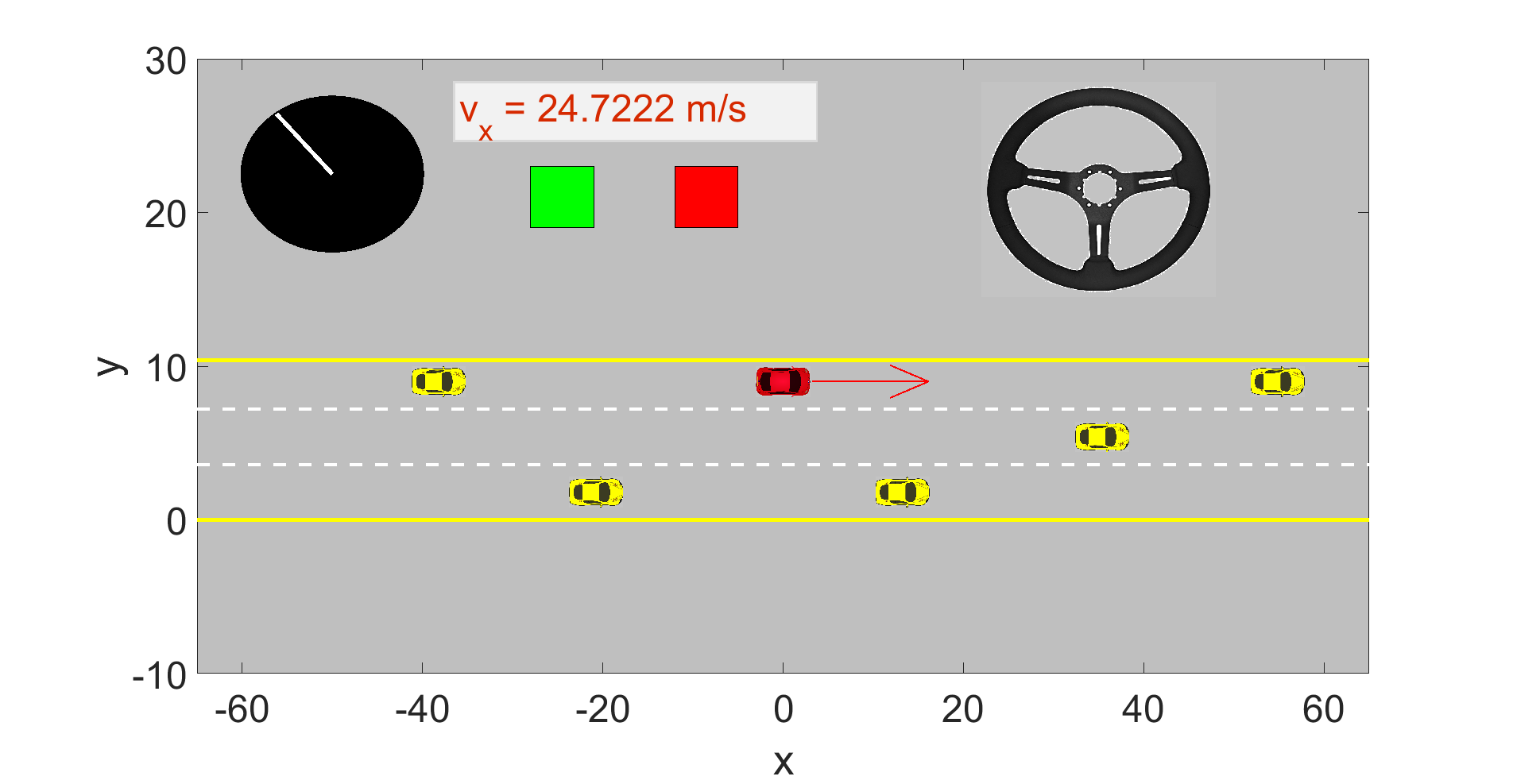,width=2.7in}}  
\put(  0,  100.0){\epsfig{file=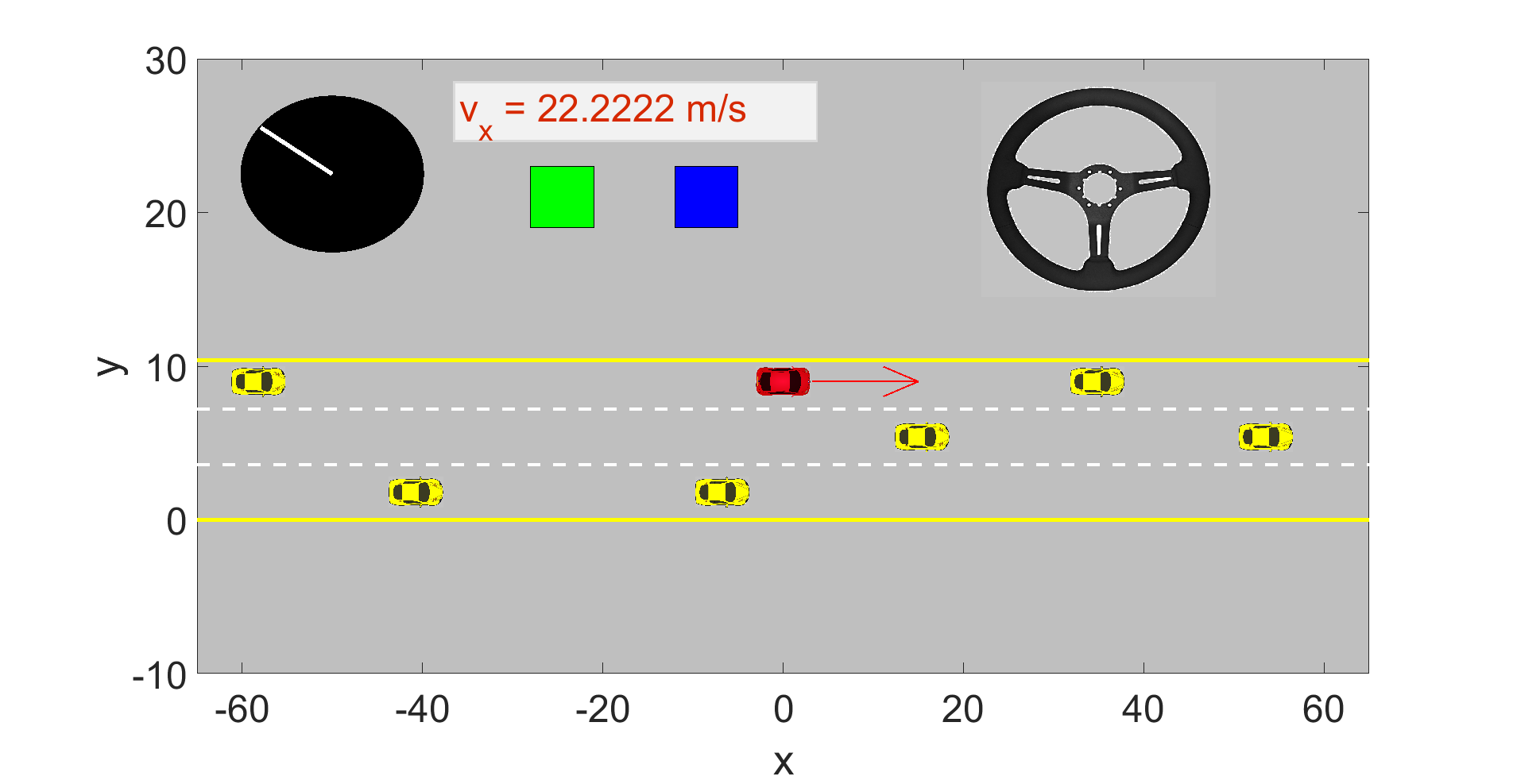,width=2.7in}}  
\put(  0,  0){\epsfig{file=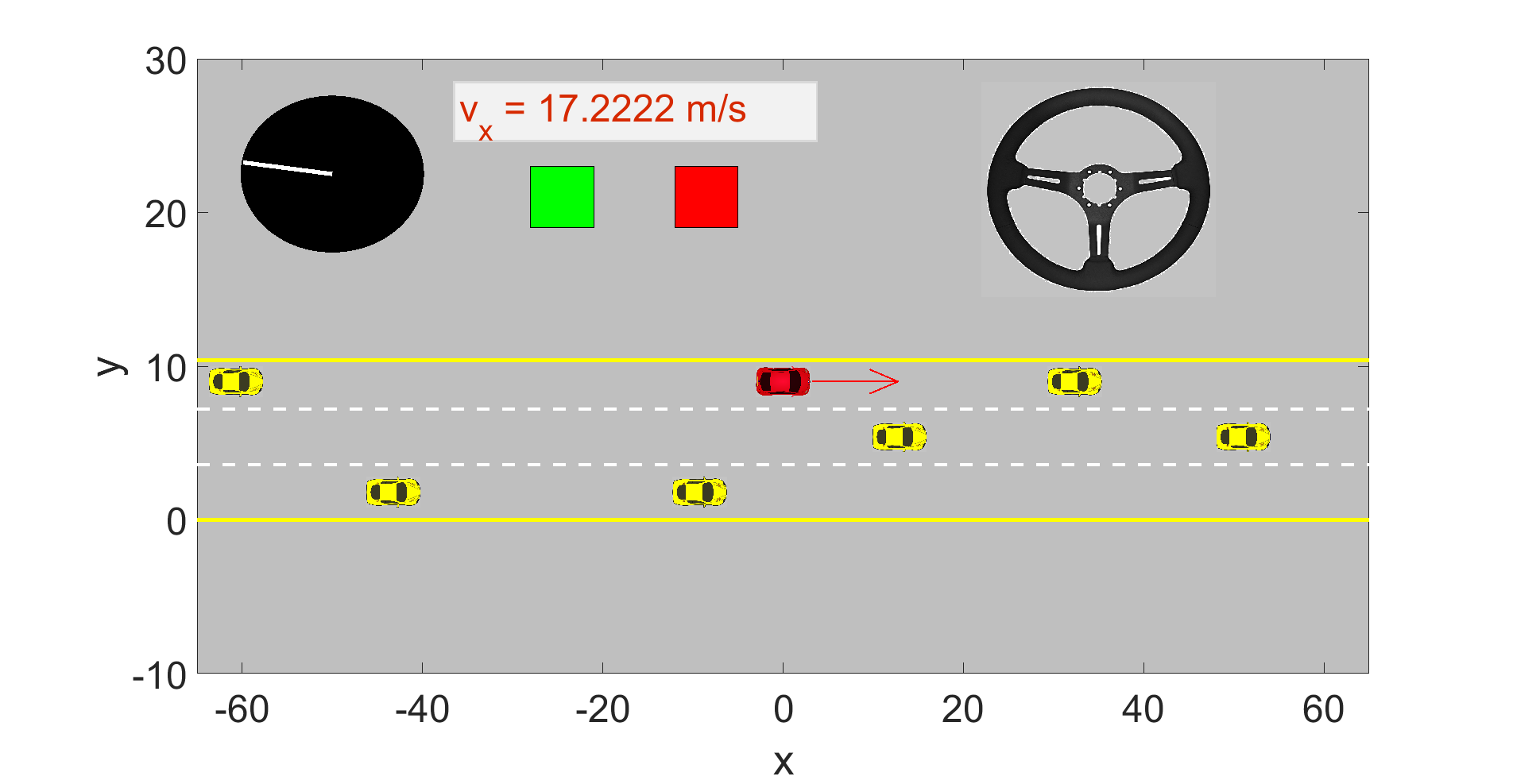,width=2.7in}}  

\put(  186,  580.0){(a)}  
\put(  186,  565.0){178[s]}  
\put(  186,  480.0){(b)}  
\put(  186,  465.0){179[s]}  
\put(  186,  380.0){(c)}  
\put(  186,  365.0){180[s]}  
\put(  186,  280.0){(d)}  
\put(  186,  265.0){181[s]}  
\put(  186,  180.0){(e)}  
\put(  186,  165.0){184[s]}  
\put(  186,  80.0){(f)}  
\put(  186,  65.0){188[s]}  

\end{picture}
\end{center}
      \caption{Level-1 vs. level-0 simulation results: Plots (a)-(f) show snapshots of the simulation at 178[s], 179[s], 180[s], 181[s], 184[s], 188[s], respectively.}
      \label{fig:Sim_Level1}
\end{figure}
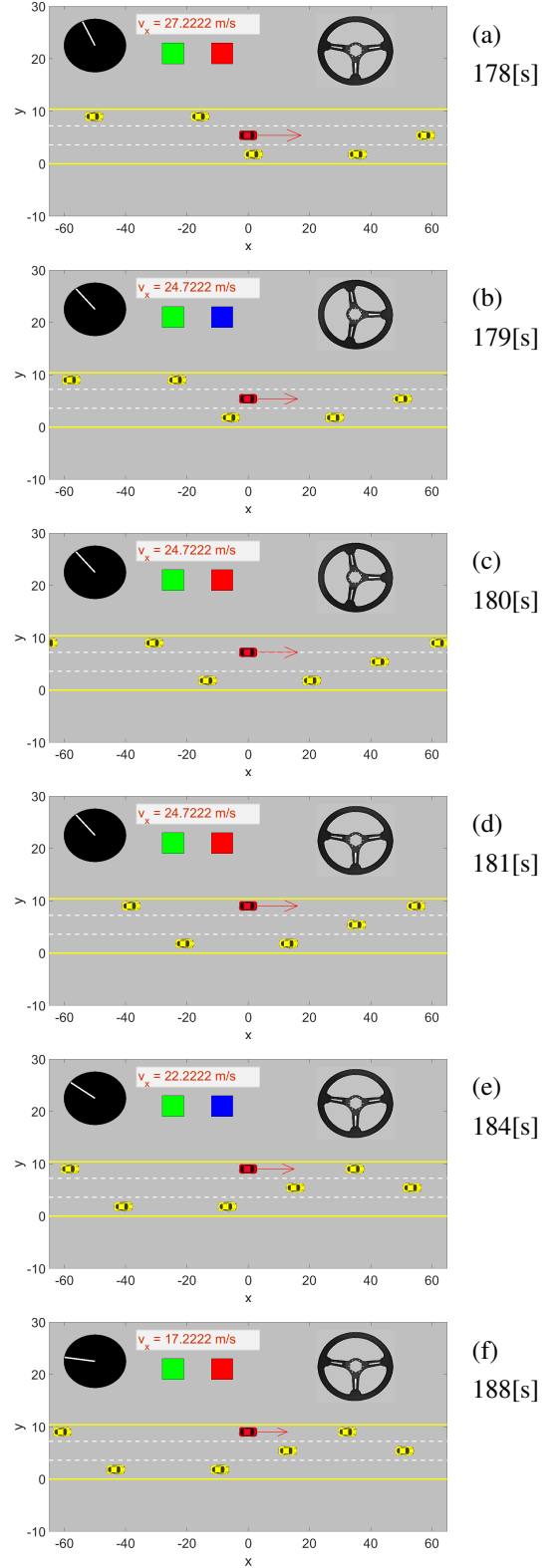

\begin{figure}[h!]
\begin{center}
\begin{picture}(220.0, 400.0)




\put(  0,  300){\epsfig{file=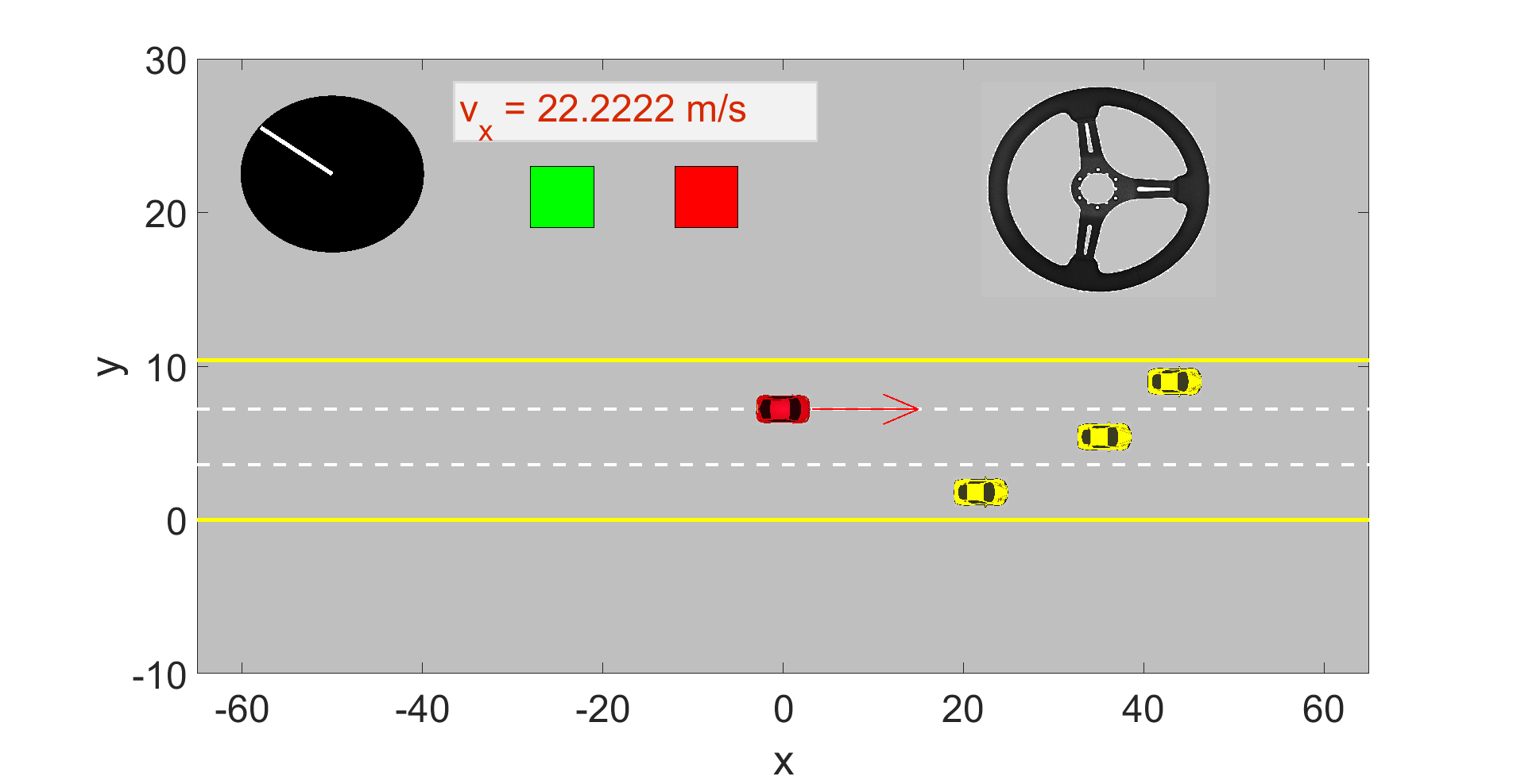,width=2.7in}}  
\put(  0,  200.0){\epsfig{file=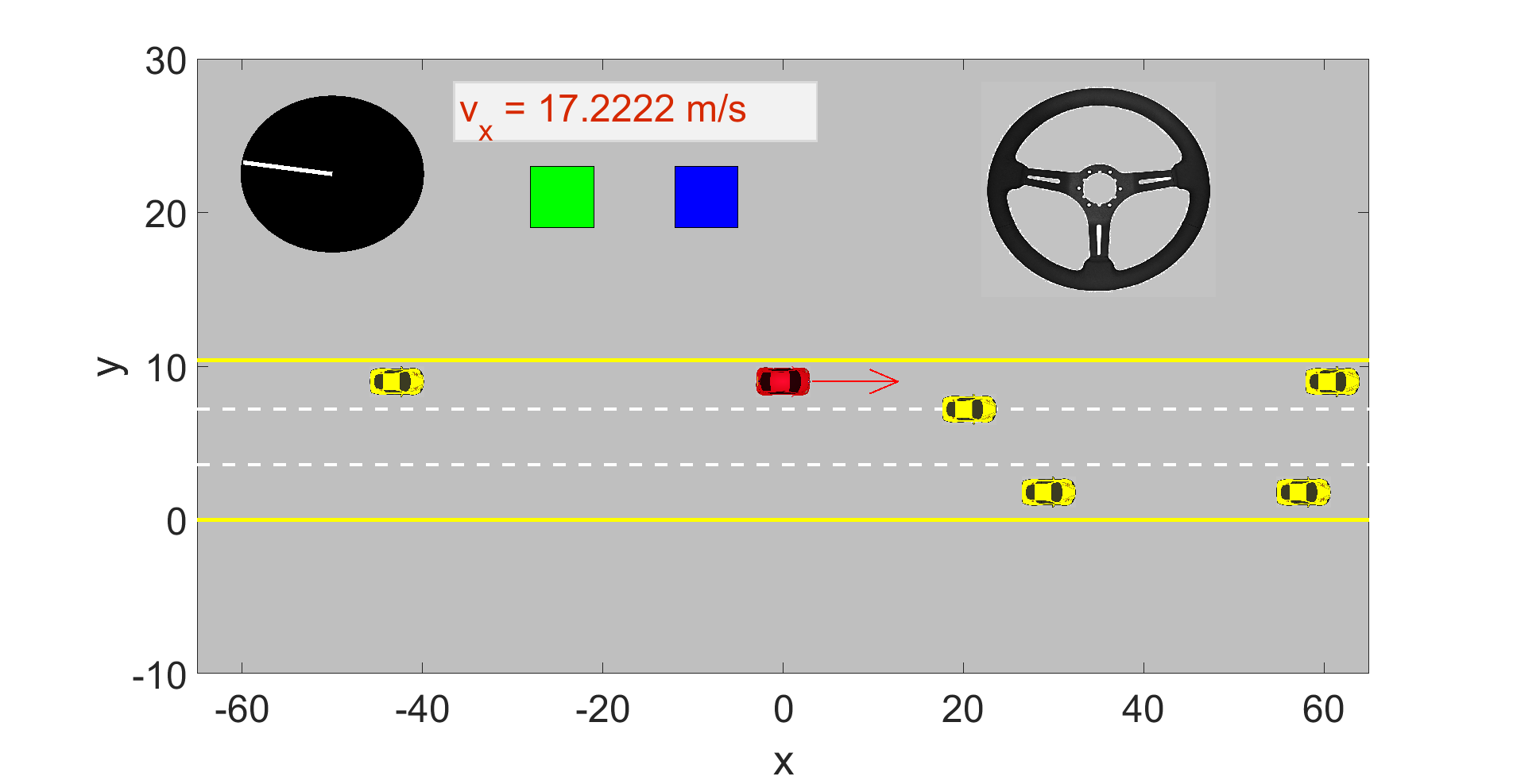,width=2.7in}}  
\put(  0,  100.0){\epsfig{file=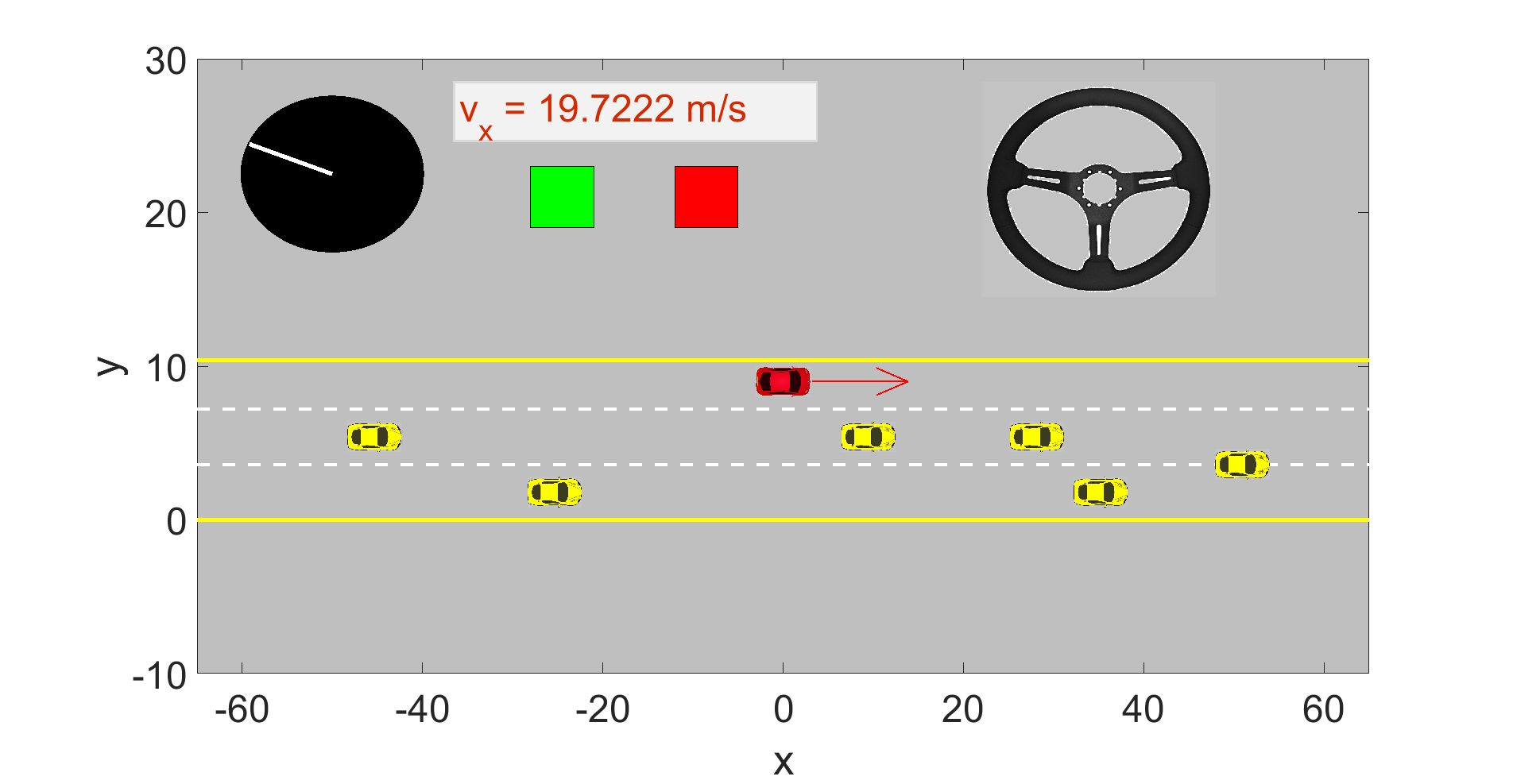,width=2.7in}}  
\put(  0,  0.0){\epsfig{file=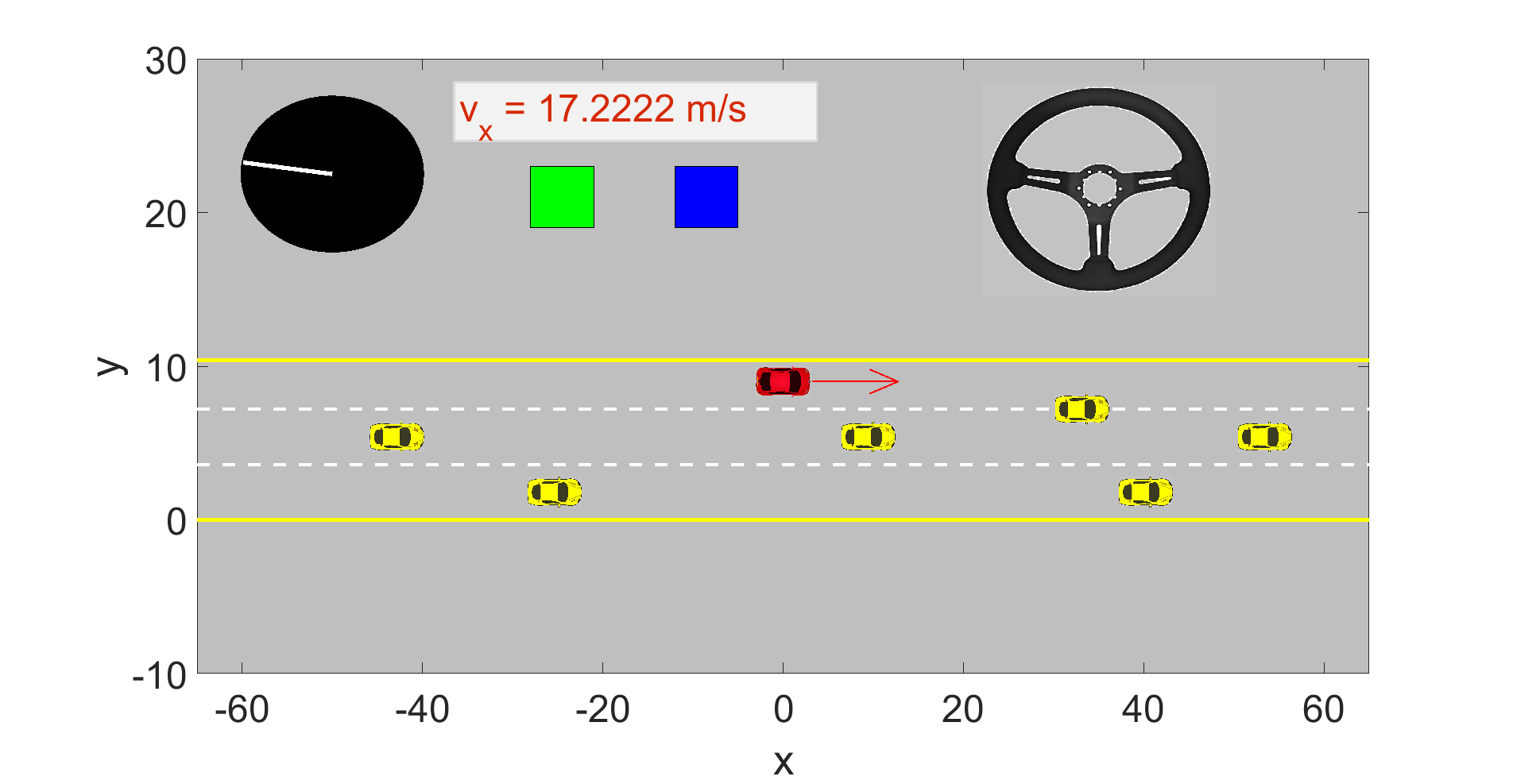,width=2.7in}}  

\put(  186,  380.0){(a)}  
\put(  186,  365.0){60[s]}  
\put(  186,  280.0){(b)}  
\put(  186,  265.0){66[s]}  
\put(  186,  180.0){(c)}  
\put(  186,  165.0){79[s]}  
\put(  186,  80.0){(d)}  
\put(  186,  65.0){80[s]}  

\end{picture}
\end{center}
      \caption{Level-2 vs. level-1 simulation results: Plots (a)-(d) show snapshots of the simulation at 60[s], 66[s], 79[s] and 80[s], respectively.}
      \label{fig:Sim_Level2}
\end{figure}
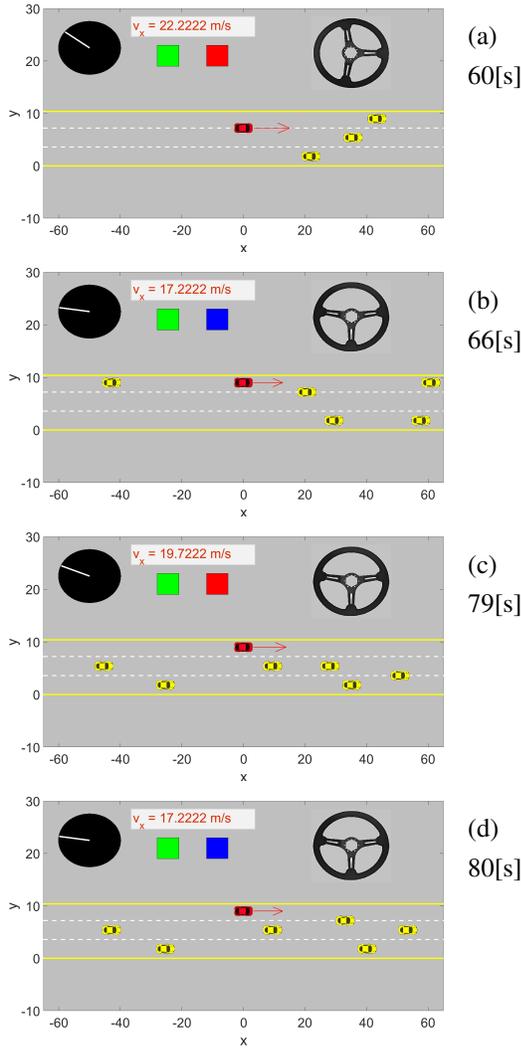

We first present simulation results to show the driving behavior of a level-1 car in a level-0 traffic environment (Fig.~\ref{fig:Sim_Level1}), and then of a level-2 car in a level-1 traffic environment (Fig.~\ref{fig:Sim_Level2}). It is noted that in both figures, the traffic is moving towards the right. The test vehicle is red and the rest of the vehicles are yellow.

In Fig.~\ref{fig:Sim_Level1}(a), the car directly in front of the test vehicle is ``approaching'' with a large relative speed and enters the ``far'' distance range. In Fig.~\ref{fig:Sim_Level1}(b), the red car decelerates and begins to steer. In Fig.~\ref{fig:Sim_Level1}(c), the red car starts to move into the lane on its left. In Fig.~\ref{fig:Sim_Level1}(d), the lane changing is completed, after 2[s]; while another car in front approaches the red car. In Fig.~\ref{fig:Sim_Level1}(e), the red car decelerates until being able to keep a stable headway, which is shown in Fig.~\ref{fig:Sim_Level1}(f). All these actions represent a reasonable driving behavior.

Similarly, Fig.~\ref{fig:Sim_Level2} shows a simulation result of a level-2 car in a level-1 traffic environment. In Fig.~\ref{fig:Sim_Level2}(a), the yellow car in the middle lane is ``approaching'' the red car, while the yellow car in the left lane is ``moving away,'' so the red car decides to change lane to the left. In Fig.~\ref{fig:Sim_Level2}(b), the yellow car in the middle lane starts to change lane to the left, so the red car needs to brake. In Fig.~\ref{fig:Sim_Level2}(c), there is no car directly in front of the red car, in which case a level-1 car would accelerate. However, because the longitudinal distance of the red car to the yellow car in the middle lane is too small and the red car has no confidence that the yellow car won't change lane to the left, in which case the red car's acceleration would lead to a distance constraint violation, the red car decides to maintain its current speed. In Fig.~\ref{fig:Sim_Level2}(d), there is a car in the traffic moving into the red car's lane, which forces the red car to decelerate.

Different driving patterns of level-1 and level-2 cars, in similar situations, reflect different levels of reasoning in decision making. It is noted that in Fig.~\ref{fig:Sim_Level2}, the cars in the environment (in yellow) also change lanes, unlike the ones in Fig.~\ref{fig:Sim_Level1}, because their actions are based on the level-1 policy, which makes them react to their own surroundings.

\begin{figure}[h!]
\begin{center}
\begin{picture}(200.0, 180.0)

\put(  0,  0){\epsfig{file=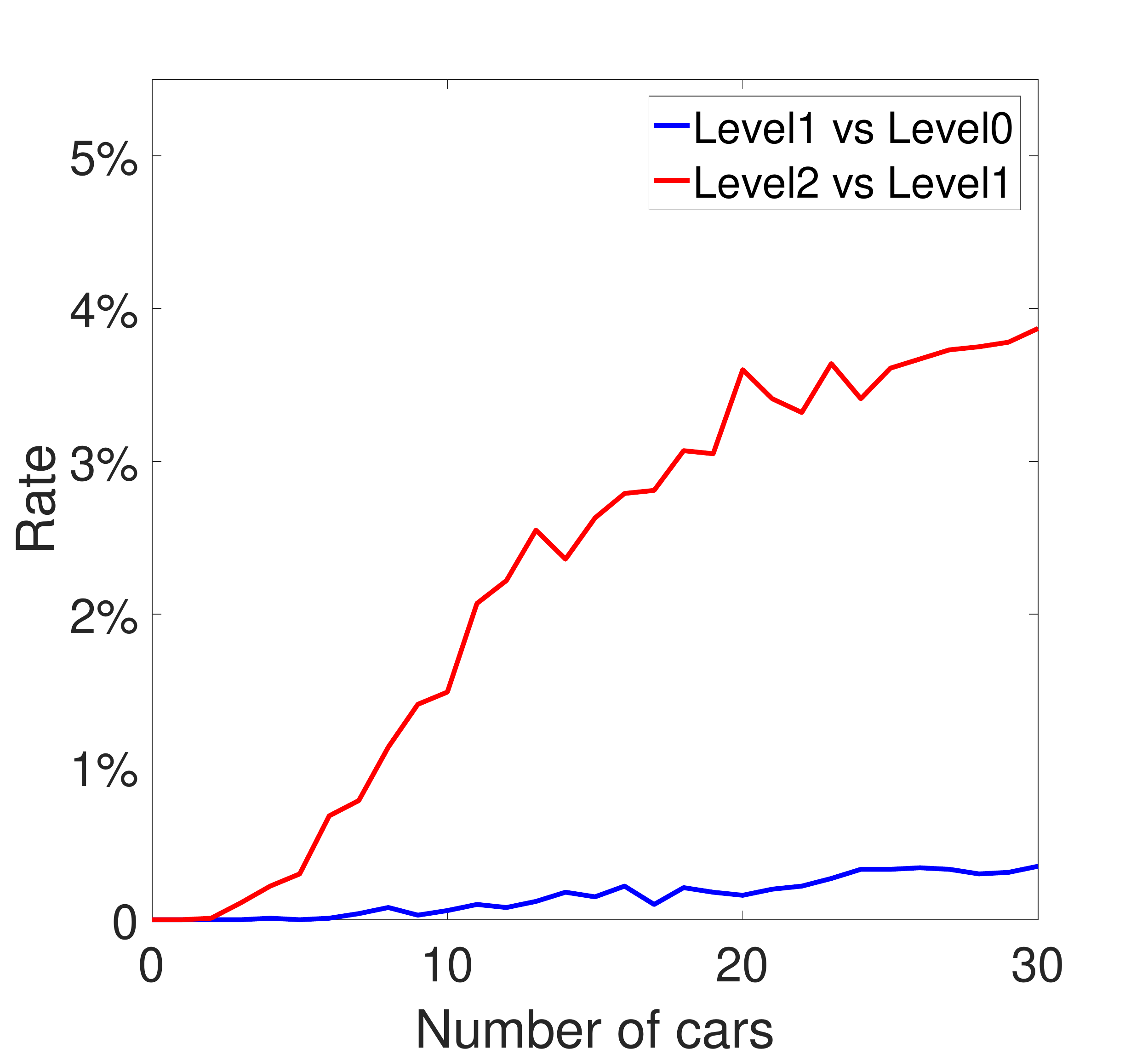,width=2.7in}}  

\end{picture}
\end{center}
      \caption{Constraint violation rates of the level-1 and level-2 policies.}
      \label{fig:colli_levelk}
\end{figure}

Fig.~\ref{fig:colli_levelk} shows the constraint violation rates of the level-1 and level-2 test cars for varying numbers of cars in the traffic. Here, ``constraint violation'' refers to the violation of the safe zone of the test car by any of the vehicles in the simulations. To obtain these rates, 10,000 simulations are run for each case (i.e., for each value of the number of cars). Each simulation lasts 200 seconds and the rates are provided as the percentage of simulation runs during which safe zones are violated. Note that the level-2 test car is placed in traffic of level-1 vehicles, and the level-1 car
is placed in the environment of level-0 vehicles. It is seen that the level-2 test car experiences higher violation rates than the level-1 test car. One explanation for this is that the traffic flow consisting of level-1 cars, where the level-2 policy is evaluated, is much harder to predict compared to the one consisting of level-0 cars, where the level-1 policy is tested.

\subsection{A comparative quantitative evaluation of Stackelberg and decision tree policies}

At first, we test the Stackelberg and the decision tree policies in a traffic consisting of only level-0 vehicles. A defining feature of the level-0 policy is that the vehicles do not change lanes. It is observed that both the Stackelberg and the decision tree policies perform well in this environment, i.e., constraint violations are not observed. This is also in agreement with the results in \cite{Yoo:12}, \cite{Yoo:13} and \cite{Claussman:15}. The figures are omitted as they provide no additional information.

\begin{figure}[h!]
\begin{center}
\begin{picture}(200.0, 180.0)

\put(  0,  0){\epsfig{file=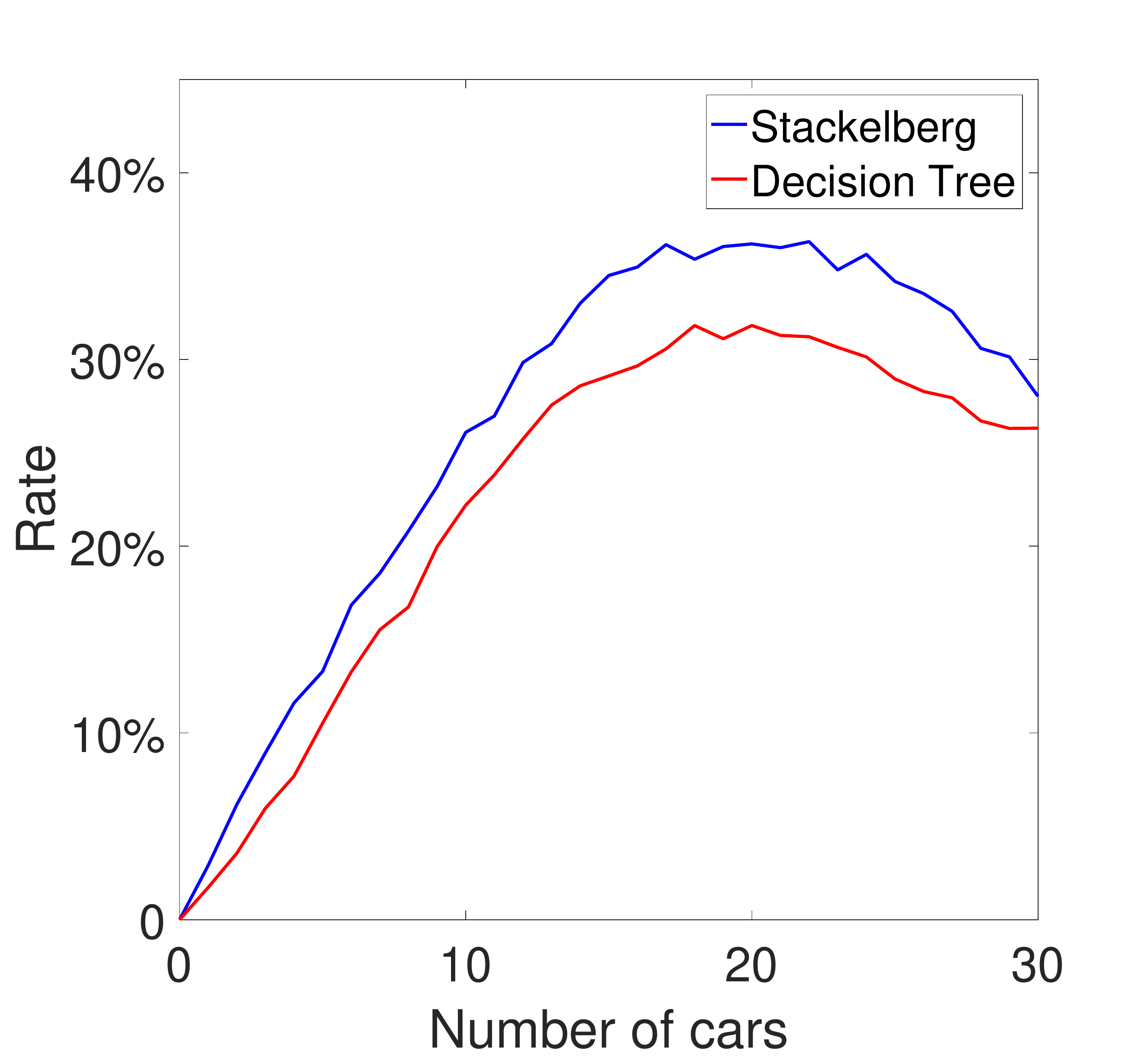,width=2.7in}}  

\end{picture}
\end{center}
      \caption{Constraint violation rates of the Stackelberg and decision tree policies.}
      \label{fig:colli_DT_SB}
\end{figure}

Next, we consider a simulated traffic environment where 10\% of the drivers make decisions based on level-0 policies, 60\% of the drivers act based on level-1 policies and 30\% use level-2 policies. These percentages of various levels are assumed based on an experimental study conducted in \cite{Costa-Gomes:09}. Fig.~\ref{fig:colli_DT_SB} shows the distance constraint violation rates for the Stackelberg and decision tree policies vs. the number of cars in the traffic. Each simulation is 200 seconds long, and 10,000 simulations are run for each case (i.e., for each value of the number of cars).

As seen in Fig.~\ref{fig:colli_DT_SB}, both approaches exhibit significant distance constraint violation percentages. Note that these violations could also be caused by our level-$k$ drivers, but as shown in Fig.~\ref{fig:colli_levelk}, compared to the numbers in Fig.~\ref{fig:colli_DT_SB}, the constraint violation rates of the level-$k$ policies are small.

The main reason for the distance constraint violations is that the developed traffic model with interacting drivers is more complex than the traffic models used for the development of the Stackelberg and decision tree algorithms. To the best of our knowledge, few works addressing the development of autonomous vehicle control policies consider traffic of similar complexity.

\begin{figure}[h!]
\begin{center}
\begin{picture}(140.0, 190.0)

\put(  17,  100){\epsfig{file=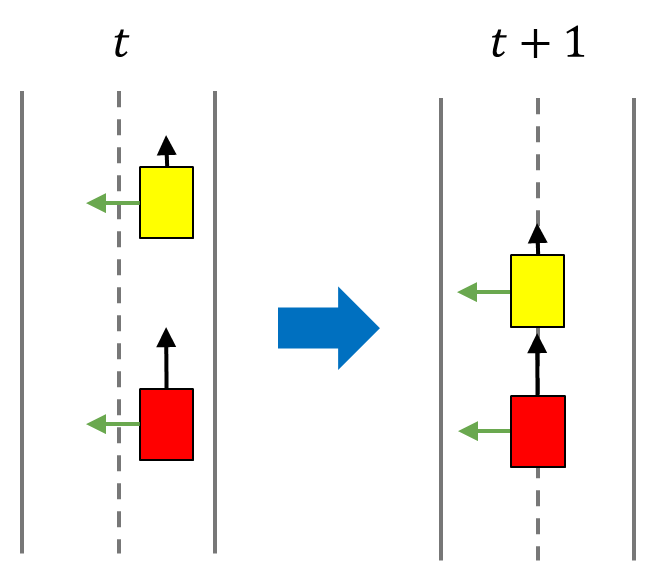,height=1.4in}}  
\put(  -1.75,  0){\epsfig{file=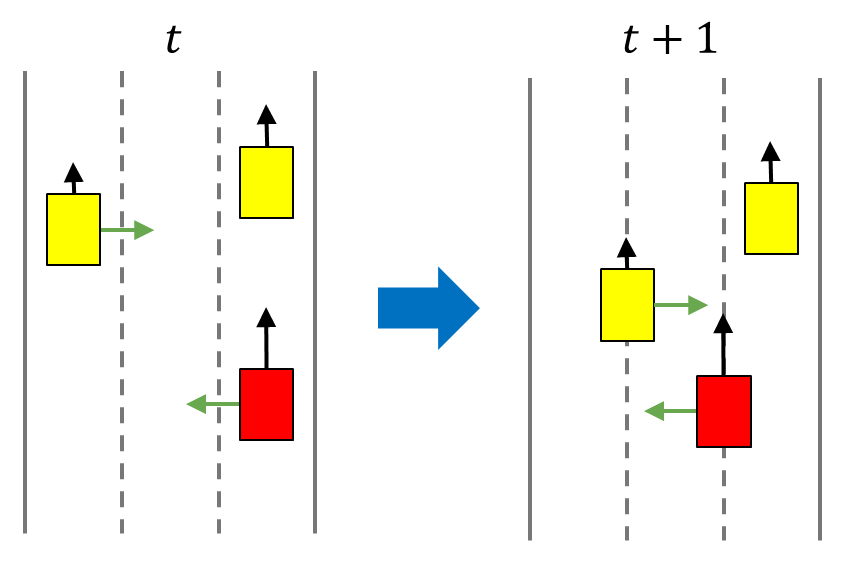,height=1.4in}}  

\put(  160,  165.0){(a)}  
\put(  160,  65.0){(b)}  

\end{picture}
\end{center}
      \caption{Scenarios for distance constraint violation.}
      \label{fig:case_violation}
\end{figure}

Fig.~\ref{fig:case_violation} shows two cases of driver interactions that are responsible for many distance constraint violations. The red rectangle indicates the car being tested, while the yellow rectangles are the cars in the traffic environment. The black arrows indicate the longitudinal velocities of the cars, while green arrows indicate the lateral velocities. The left hand side presents the scenarios at time $t$, while the right hand side presents the scenarios at time $t+1$. In Fig.~\ref{fig:case_violation}(a), the red car is starting to change lane to the left, trying to overtake the front car, while at the same time the front car is also starting to change lane because of some other car ahead of it (not shown in this figure). As a result, both cars are changing lanes while their longitudinal distance keeps decreasing until below the safe distance constraint. Although the red car may brake hard after the lane change, trying to avoid a distance constraint violation, at that time its range is already too small for it to avoid that by braking. In Fig.~\ref{fig:case_violation}(b), the red car is starting to change lane to the left to overtake the car in front of it. Although the car in front remains in its own lane, there is some other car in the leftmost lane starting to move into the middle lane as well. As a result, both the red car and the yellow car in the leftmost lane are changing lanes to the middle and eventually violate the safe distance between them.

We note that challenging scenarios as the ones above can greatly facilitate the testing of autonomous driving control algorithms. In fact, the above two cases are also dangerous situations for a human driver. Many traffic accidents result from the driver misjudging the potential actions of the surrounding vehicles.

\begin{figure}[h!]
\begin{center}
\begin{picture}(160.0, 100.0)

\put(  20,  6.8){\epsfig{file=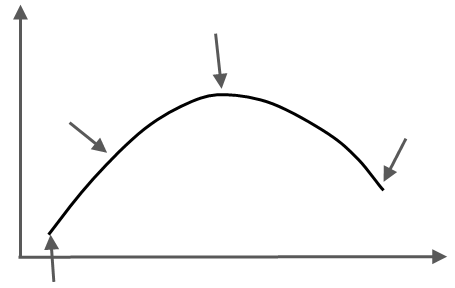,width=1.6in}}  
\small

\put(  17,  82){Rate}  
\put(  80,  2){Traffic density}  
\put(  29,  -0.5){\textcircled{1}}  
\put(  27,  50){\textcircled{2}}  
\put(  70,  75){\textcircled{3}}  
\put(  123,  49){\textcircled{4}}  

\end{picture}
\end{center}
      \caption{Diagram of constraint violation rate.}
      \label{fig:FDT_CVR}
\end{figure}
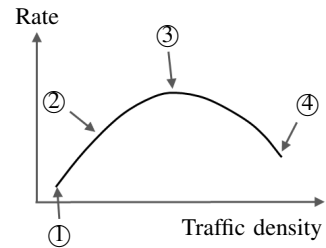

Note also that the constraint violation rate first increases and then decreases as the number of cars in the environment, which reflects the traffic density, increases. As Fig.~\ref{fig:FDT_CVR} shows, when the traffic is very sparse, cars on the road can drive almost freely and have low chance of having a constraint violation (\textcircled{1}). As the number of cars in the traffic increases, the chance of experiencing an incident also increases (\textcircled{2}) (until a peak \textcircled{3}). When the traffic becomes very dense, for instance, in a traffic jam, the average traveling speed becomes low, and at the same time each car mostly stays in its own lane and has few lane change maneuvers. As a result, the probability of having constraint violations becomes low again (\textcircled{4}).

\begin{figure}[h!]
\begin{center}
\begin{picture}(200.0, 180.0)

\put(  0,  0){\epsfig{file=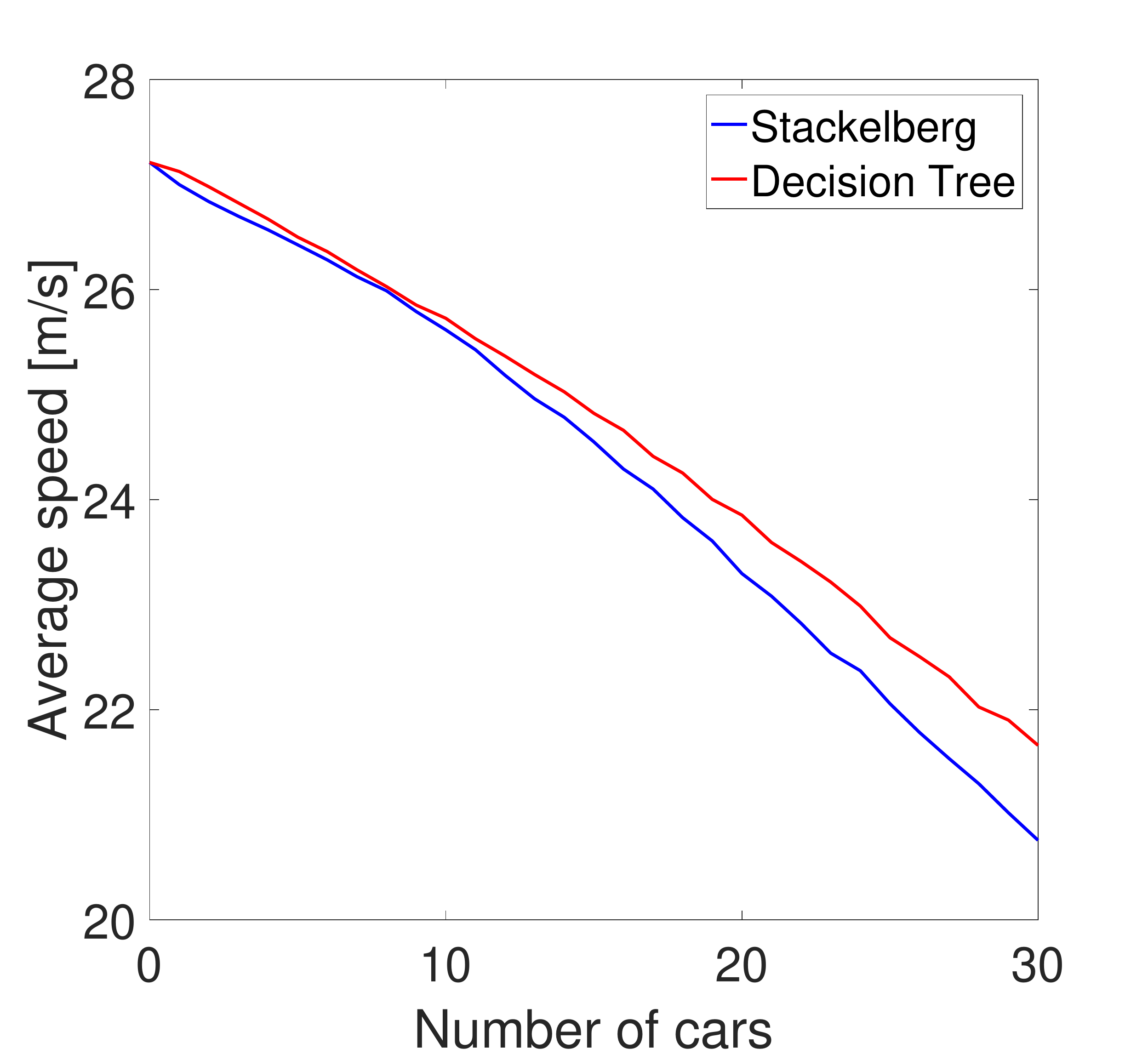,width=2.7in}}  

\end{picture}
\end{center}
      \caption{Average traveling speed of the Stackelberg and decision tree policies.}
      \label{fig:speed_DT_SB}
\end{figure}

Apart from using the constraint violation rate as a metric to measure the safety and robustness of the Stackelberg and decision tree policies, we also use the average travel speed to measure the driving performance. It can be observed from Fig.~\ref{fig:speed_DT_SB} that the decision tree policy has better driving performance compared to the Stackelberg policy.

\begin{figure}[h!]
\begin{center}
\begin{picture}(200.0, 180.0)

\put(  0,  0){\epsfig{file=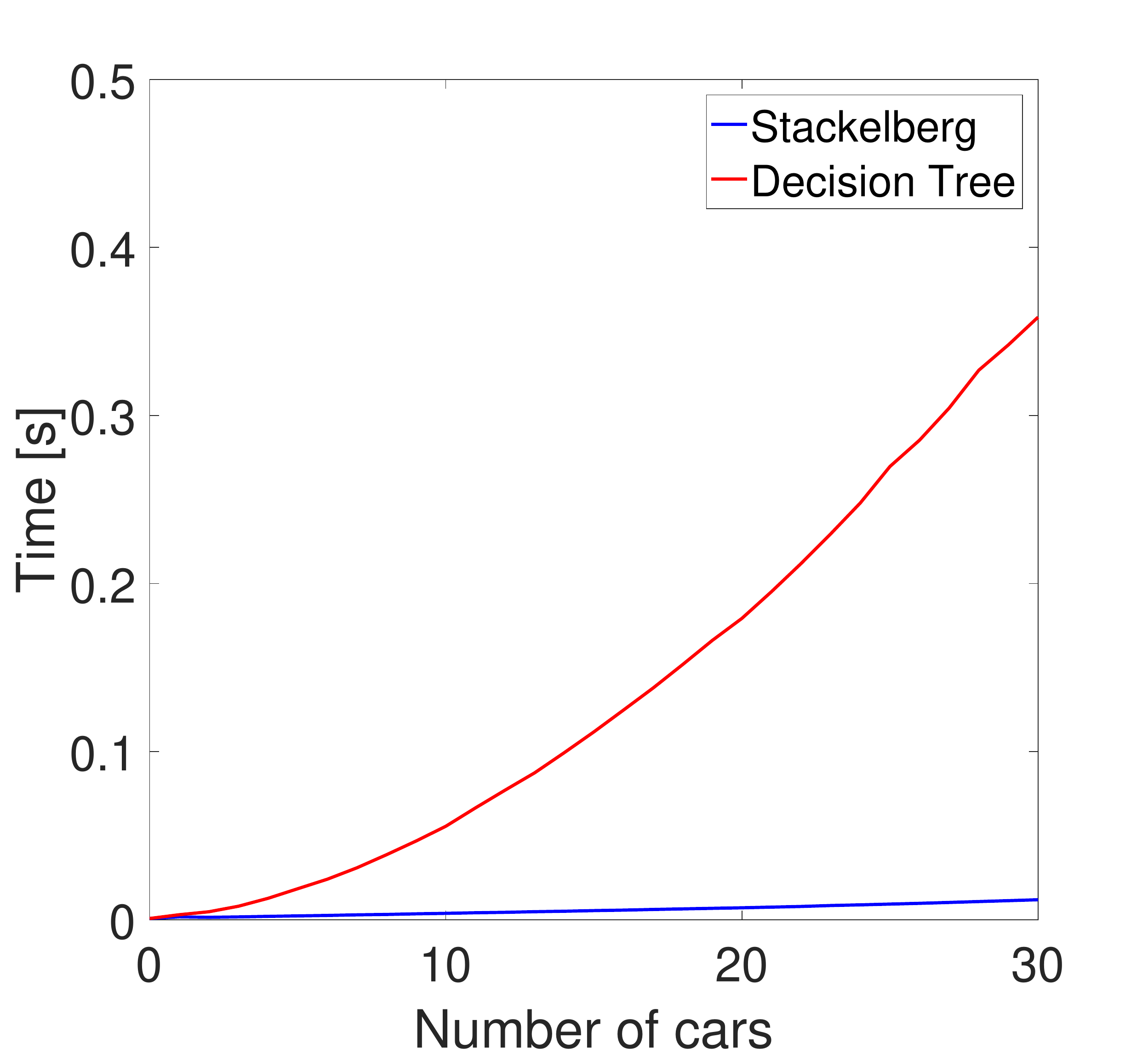,width=2.7in}}  

\end{picture}
\end{center}
      \caption{Computational time of the Stackelberg and decision tree policies.}
      \label{fig:computationT_DT_SB}
\end{figure}

Fig.~\ref{fig:computationT_DT_SB} compares the computational load of the two autonomous driving control methods. The numbers shown in the plot are the average time consumed to run the 200[s] simulation. Because the decision tree policy needs to evaluate 49 profiles of 2-layer actions in total, its required computation load is higher than that of the Stackelberg policy. As a summary, our implementation of the decision tree algorithm is better than our implementation of the Stackelberg algorithm in terms of both safety and performance, while the price paid is the higher computational cost. The numbers in Fig.~\ref{fig:computationT_DT_SB} are obtained using the Java $System.nanoTime()$ function, and the simulations are run on a desktop with i7-4790 3.60GHz CPU and with Eclipse Java Neon platform. Note also that as Fig.~\ref{fig:computationT_DT_SB} indicates the simulator is quite fast.


\subsection{Optimal autonomous driving controller calibration}
One of the potential uses of the proposed traffic modeling approach is for calibrating parameter values in the autonomous driving control algorithms. We illustrate this using the decision tree policy as an example.

The optimization objective is defined by considering both safety and performance: the goal is to maximize the following reward function:
\begin{equation}
R_{\rm obj}= p_1 (-\bar{c}) + p_2 \frac{\bar{v}_x-v_{\min}}{v_{\max}-v_{\min}},
\end{equation}
where the weights $p_1$ and $p_2$ are determined by the user, $\bar{c}$ is the constraint violation rate defined as in Fig.~\ref{fig:colli_DT_SB}, $\bar{v}_x$ is the average speed during the 200[s]-simulations, and $v_{\min}$ and $v_{\max}$ are the lower and the upper bounds of the test vehicle's speed, respectively. Note that this reward function is designed such that each of its terms is dimensionless. The parameters optimized in this example are the ratio $\frac{w_{l1}}{w_{l2}}$, representing the weighting of the two layers in the evaluation metric function \eqref{eq:Rtot} in the decision tree evaluations, and $x_B$, the size of region B in the longitudinal direction -- a threshold of triggering the path planner. These two parameters are selected for optimization since they have indirect influence on safety and performance, making them difficult to set from intuition. Note that the influence of other parameters, for example, $w_1,\cdots,w_4$, in the decision tree evaluation metric function, is more intuitive and they can be tuned more easily.

\begin{figure}[h!]
\begin{center}
\begin{picture}(400.0, 260.0)

\put(  -12,  130){\epsfig{file=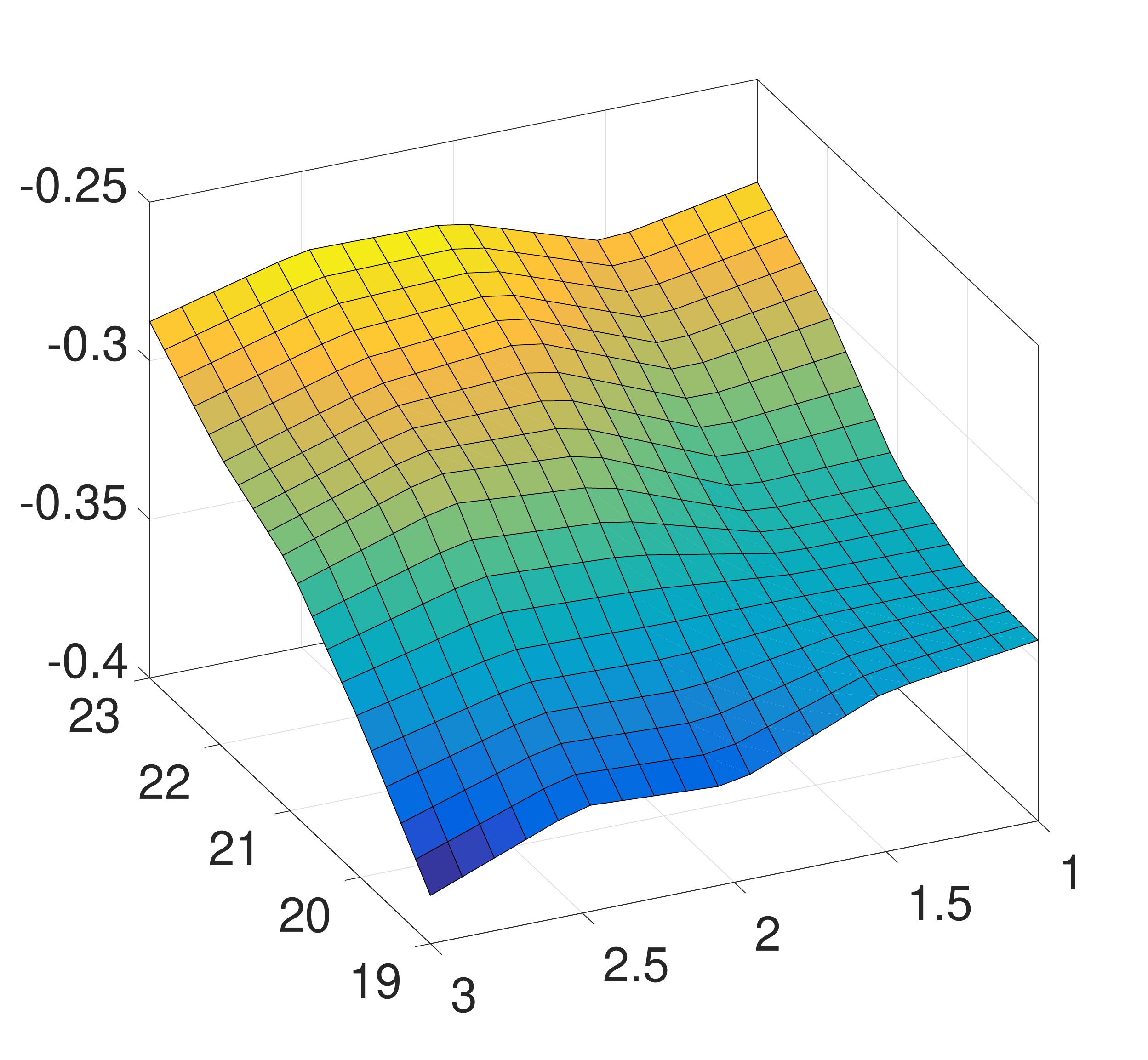,width=2in}}  
\put(  125,  130){\epsfig{file=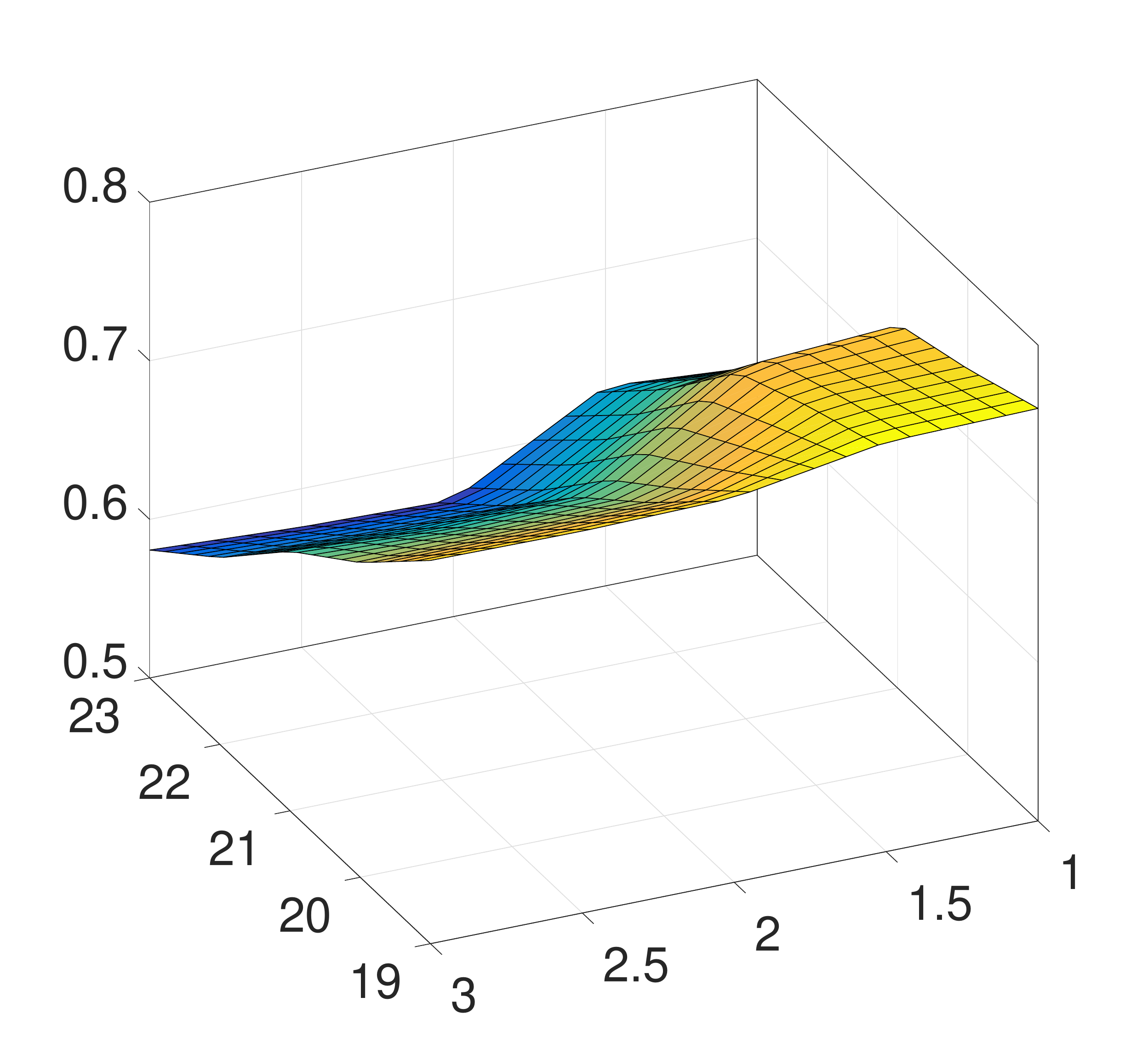,width=2in}}  

\put(  -12,  0){\epsfig{file=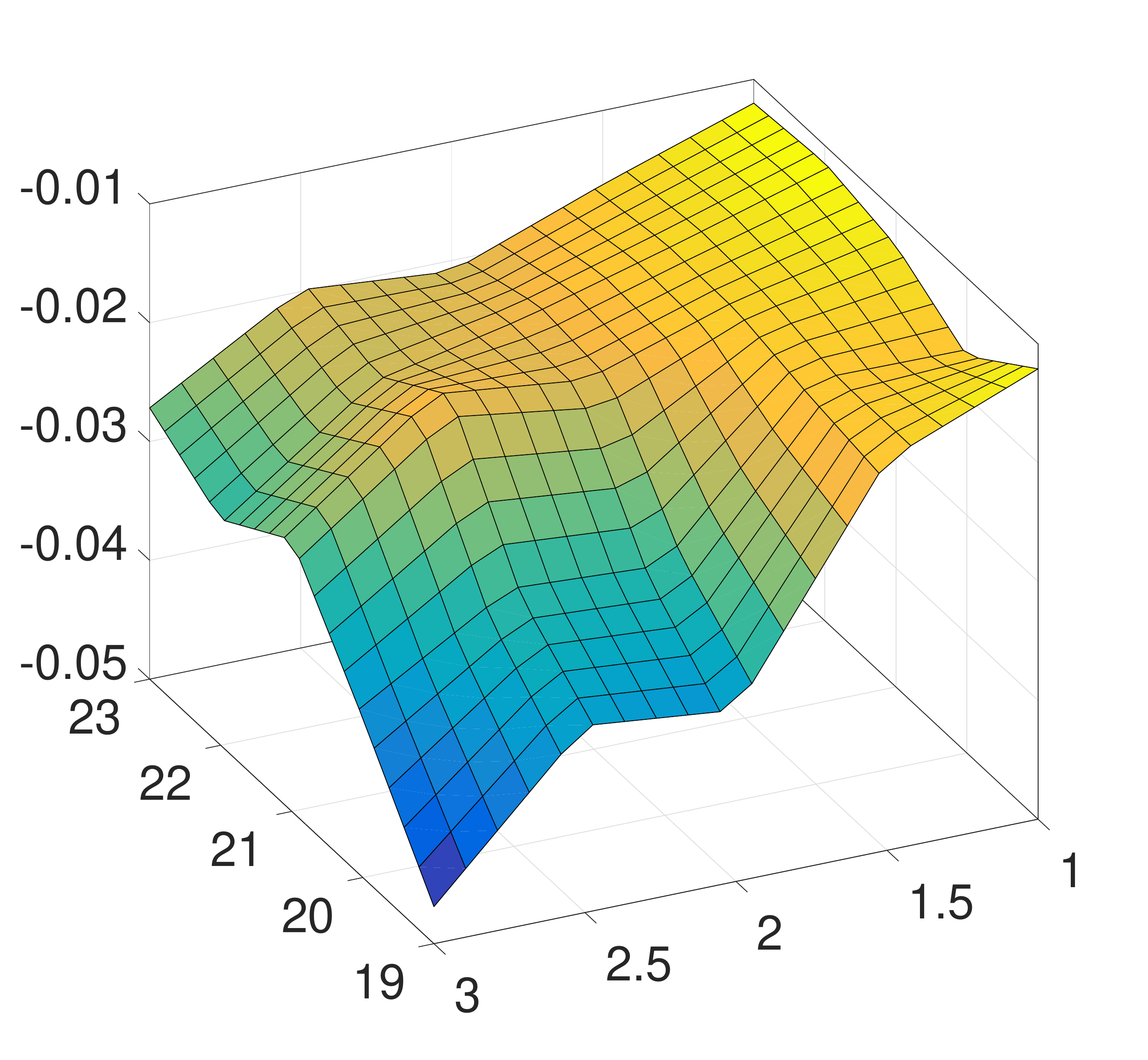,width=2in}}  
\put(  125,  0){\epsfig{file=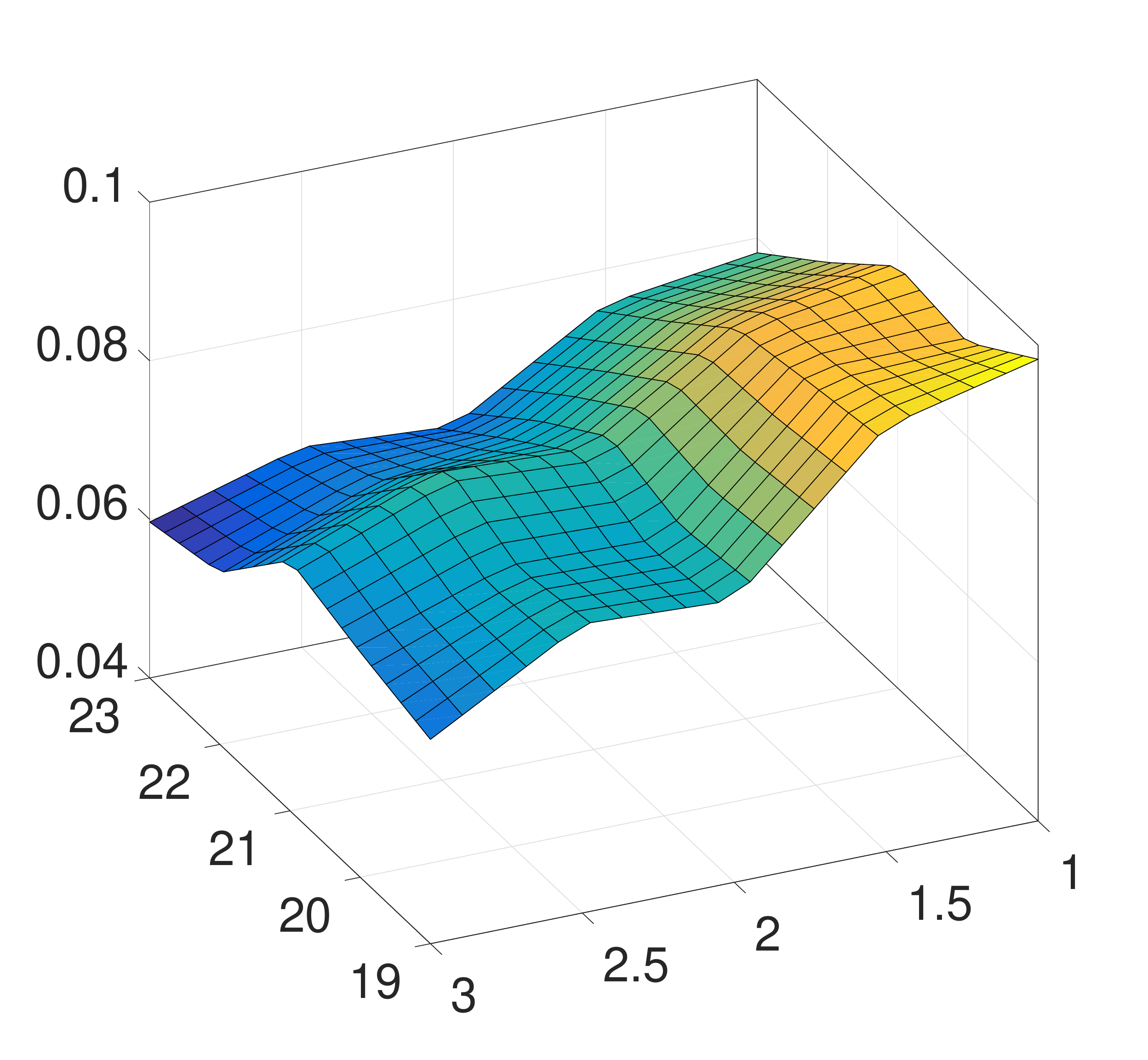,width=2in}}  

\put(  100,  248.0){(a)}  
\put(  240,  248.0){(b)}  
\put(  100,  118.0){(c)}  
\put(  240,  118.0){(d)}  

\put(  90,  136.0){$\frac{w_{l1}}{w_{l2}}$}  
\put(  5,  142.0){$x_B$}  
\put(  228,  136.0){$\frac{w_{l1}}{w_{l2}}$}  
\put(  142,  142.0){$x_B$}  
\put(  90,  6.0){$\frac{w_{l1}}{w_{l2}}$}  
\put(  5,  12.0){$x_B$}  
\put(  228,  6.0){$\frac{w_{l1}}{w_{l2}}$}  
\put(  142,  12.0){$x_B$}  

\put(  10,   248.0){$R_{\rm obj}$}  
\put(  145,  248.0){$R_{\rm obj}$}  
\put(  10,  118.0){$R_{\rm obj}$}  
\put(  145,  118.0){$R_{\rm obj}$}  

\end{picture}
\end{center}
      \caption{Parameter optimization results corresponding to different reward function designs. (a) $p_1$=1, $p_2$=0, (b) $p_1$=0, $p_2$=1, (c) $p_1$=0.7, $p_2$=0.3, (d) $p_1$=0.6, $p_2$=0.4.}
      \label{fig:Opti_surf}
\end{figure}
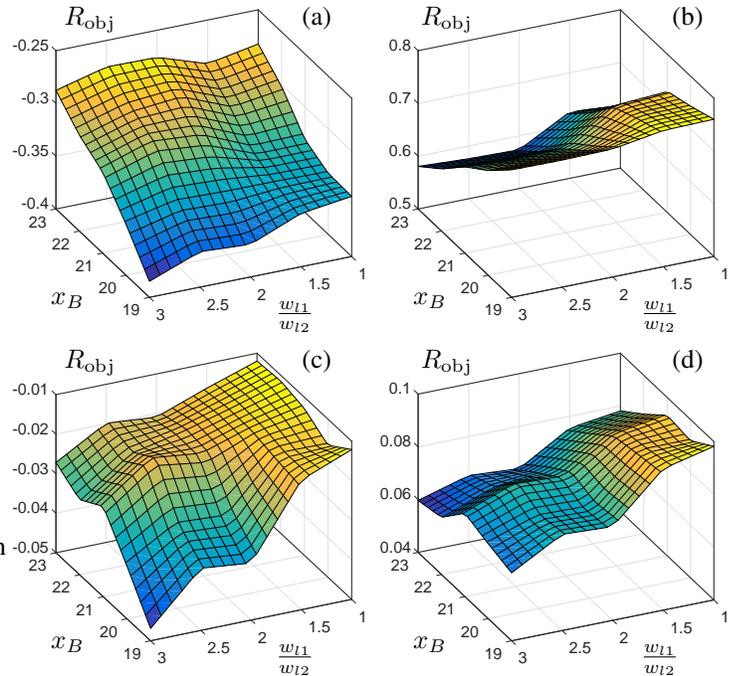

Fig.~\ref{fig:Opti_surf} shows the surface of the reward values as a function of $\frac{w_{l1}}{w_{l2}}$, $x_B$ corresponding to different weight selections $p_1$, $p_2$, in the presence of a 20-car traffic. These figures can be used to pick the best pair of $(\frac{w_{l1}}{w_{l2}},x_B)$ for a given reward function. For example, for maximum safety ($p_1=1$, $p_2=0)$, Fig.~\ref{fig:Opti_surf}(a) shows that the best pair is $(2.5,23)$. The rate of constraint violation with this pair is $27.5\%$; while for the original selection $(2,21)$ in the previous section, the corresponding violation rate is $31.8\%$.

\section{Summary}\label{sec:Summary}
In this paper, a hierarchical reasoning game theory based approach to model interacting driver behavior in traffic was presented. The proposed method provides an approach to simulate interactive driver behavior under the given traffic conditions.

A traffic simulator was developed using level-$k$ driver models. It can be used for testing and verification of autonomous driving algorithms, and for discovery of challenging trajectories and scenarios that can facilitate the testing of future autonomous vehicles. To illustrate the simulator use, we have defined and tested two autonomous vehicle control policies in terms of safety and performance. Our traffic simulator can also be used for parameter calibration of these policies by a simulation-based optimization approach.

\ifCLASSOPTIONcaptionsoff
  \newpage
\fi



%
%
%

\section*{}\bibliographystyle{IEEEtran}
\bibliography{ref_6Apr2016}

%








\end{document}